\documentclass[journal]{IEEEtran}
\usepackage[square,numbers,sort&compress]{natbib}

\ifCLASSINFOpdf
\else
\fi
\usepackage{epsfig}
\usepackage{graphicx}
\usepackage{subfigure}

\usepackage[cmex10]{amsmath}
\usepackage{mathrsfs}
\usepackage{amsmath}
\usepackage{float}
\usepackage{algorithm}
\usepackage{amsthm}
\usepackage{multirow}
\usepackage{xcolor}

\usepackage{color, soul}
\usepackage{algpseudocode}
\DeclareMathOperator*{\argmin}{argmin}

\begin {document}
%
\title{Exploring Representativeness and Informativeness for Active Learning}
%
%
%

\author{Bo~Du,~
        Zengmao~Wang,~
        Lefei~Zhang,~
        Liangpei~Zhang,~
        Wei~Liu,~
        Jialie~Shen,~
        and~Dacheng~Tao,~
\thanks{Manuscript received June 3, 2015; revised August 21, 2015; accepted October 27, 2015. This work was supported in part by
the National Basic Research Program of China (973 Program) under Grant 2012CB719905,
by the National Natural Science
Foundation of China under Grants 61471274, 41431175 and 61401317, by the
Natural Science Foundation of Hubei Province under Grant 2014CFB193, by the Fundamental Research Funds for the Central Universities under Grant 2042014kf0239, and by Australian Research Council Projects DP-140102164 and FT-130101457.}
\thanks{B. Du, Z. Wang, and L. Zhang are with the State Key Laboratory of Software Engineering, School of Computer,  Wuhan University, China e-mail:(gunspace@163.com, wzm902009@gmail.com, zhanglefei@ieee.org), and L. Zhang is also with  Department of Computing, Hong Kong Polytechnic University, Kowloon, HK.}
\thanks{L. Zhang is with  the Collaborative Innovation Center of Geospatial Technology, State Key Laboratory of Information Engineering in Surveying, Mapping and Remote Sensing, Wuhan University, China e-mail:(zlp62@whu.edu.cn). W. Liu is with IBM T. J. Watson  Research Center, Yorktown Heights NY 10598 USA e-mail:(wliu@ee.columbia.edu). J. Shen is with School of Information  Systems, Singapore Management University, Singapore e-mail:(jlshen@smu.edu.sg). D. Tao is with the QCIS, University of  Technology, Sydney NSW 2007, Australia e-mail:(dacheng.tao@gmail.com).}
\thanks{\copyright 20XX IEEE. Personal use of this material is permitted. Permission from IEEE must be obtained for all other uses, in any current or future media, including reprinting/republishing this material for advertising or promotional purposes, creating new collective works, for resale or redistribution to servers or lists, or reuse of any copyrighted component of this work in other works.}
}%

%
%

%

\maketitle

\begin{abstract}
How can we find a general way to choose the most suitable samples for training a classifier? Even with very limited prior information? Active learning, which can be regarded as an iterative optimization procedure, plays a key role to construct a refined training set to improve the classification performance in a variety of applications, such as text analysis, image recognition, social network modeling, \textit{etc.}

Although combining representativeness and informativeness of samples has been proven promising for active sampling, state-of-the-art methods perform well under certain data structures. Then can we find a way to fuse the two active sampling criteria without any assumption on data? This paper proposes a general active learning framework that effectively fuses the two criteria. Inspired by a two-sample discrepancy problem, triple measures are elaborately designed to guarantee that the query samples not only possess the representativeness of the unlabeled data but also reveal the diversity of the labeled data. Any appropriate similarity measure can be employed to construct the triple measures. Meanwhile, an uncertain measure is leveraged to generate the informativeness criterion, which can be carried out in different ways.

Rooted in this framework, a practical active learning algorithm is proposed, which exploits a radial basis function together with the estimated probabilities to construct the triple measures and a modified Best-versus-Second-Best strategy to construct the uncertain measure, respectively. Experimental results on benchmark datasets demonstrate that our algorithm consistently achieves superior performance over the state-of-the-art active learning algorithms.
\end{abstract}


\IEEEpeerreviewmaketitle
\section{Introduction}
\IEEEPARstart{T}{he} past decades have witnessed a rapid development of cheaply collecting huge data, providing the opportunities of intelligently classifying data using machine learning techniques\cite{T1,T2,T3,T4,T5,T6}. In classification tasks, a sufficient amount of labeled data is obliged to be provided to a classification model in order to achieve satisfactory classification accuracy\cite{T7,T8,T9,T10,T12}. However, annotating such an amount of data manually is time consuming and sometimes expensive. Hence it is wise to select fewer yet informative samples for labeling from a pool of unlabeled samples, so that a classification model trained with these optimally chosen samples can perform well on unseen data samples. If we select the unlabeled samples randomly, there would be redundancy and some samples may bias the classification model, which will eventually result in a poor generalization ability of the model. Active learning methodologies address such a challenge by querying the most informative samples for class assignments \cite{B2009,He2010,LC2011}, and the informativeness criterion for active sampling has been successfully applied to many data mining and machine learning tasks\cite{RW2013,SK2013,MJ2013,JN2014,C3,C4,C5}. Although active learning has been developed based on many approaches\cite{HW2009,JZ2015,XZ2010,OM2014}, the dream to query the most informative samples is never changing\cite{EE2013,C1,C2}.

Essentially, active learning is an iterative sampling + labeling procedure. At each iteration, it selects one sample for manually labeling, which is expected to improve the performance of the current classifier \cite{MX2012,EG2013}. Generally speaking, there are two main sampling criteria in designing an effective active learning algorithm, that is, informativeness and representativeness \cite{SR2014}. Informativeness represents the ability of a sample to reduce the generalization error of the adopted classification model, and ensures less uncertainty of the classification model in the next iteration. Representativeness decides whether a sample can exploit the structure underlying unlabeled data \cite{B2009}, and many applications have pay much attention on such information\cite{L2013,Lu2014,Lu12013,Lu32013,Lu42013}. Most popular active learning algorithms deploy only one criterion to query the most desired samples. The approaches drawing on informativeness attracted more attention in the early research of active learning. Typical approaches include: 1) query-by-committee, in which several distinct classifiers are used, and the samples are selected with the largest disagreements in the labels predicted by these classifiers \cite{HM1992,PR2004,RS2002}; 2) max-margin sampling, where the samples are selected according to the maximum uncertainty via the distances to the classification boundaries \cite{MX2011,SD2007,SD2001}; 3) max-entropy sampling, which uses entropy as the uncertainty measure via probabilistic modeling \cite{JM2012,YR2007,AF2009,AP2012}. The common issue of the above active learning methods is that they may not be able to take full advantage of the information of abundant unlabeled data, and query the samples merely relying on scarce labeled data. Therefore, they may be prone to a sampling bias.

Hence, a number of active learning algorithms have recently been proposed based on the representativeness criterion to exploit the structure of unlabeled data in order to overcome the deficiency of the informativeness criterion. Among these methods, there are two typical means to explore the representativeness in unlabeled data. One is clustering methods~\cite{SD2008,HA2004}, which exploit the clustering structure of unlabeled data and choose the query samples closest to the cluster centers. The performance of clustering based methods depends on how well the clustering structure can represent the entire data structure. The other is optimal experimental design methods~\cite{RZ2012,PM2005,KJ2006}, which try to query the representative examples in a transductive manner. The major problem of experimental design based methods is that a large number of samples need to be accessed before the optimal decision boundary is found, while the informativeness of the query samples is almost ignored.

Since either single criterion cannot perform perfectly, it is natural to combine the two criteria to query the desired samples. Wang and Ye \cite{ZJ2013} introduced an empirical risk minimization principle to active learning, and derived an empirical upper-bound for forecasting the active learning risk. By doing so, the discriminative and representative samples are effectively queried, and the uncertainty measure stems from a hypothetical classification model (\textit{i.e.}, a regression model in a kernel space). However, a bias would be caused if the hypothetical model and the true model differed in some aspects. Huang \textit{et al.} \cite{SR2014} also proposed an active learning framework combining informativeness and representativeness, in which classification uncertainty is used as the informativeness measurement and a semi-supervised learning algorithm is introduced to discover the representativeness of unlabeled data\cite{MF2010,SR2008}. To run the semi-supervised learning algorithm, the input data structure should satisfy the semi-supervised assumption to guarantee the performance\cite{OB2006}.

As reviewed above, we argue that the current attempts to fuse the informativeness and representativeness in active learning may be susceptible to certain assumptions and constraints on input data. Hence, can we find a general way to combine the two criteria in design active learning methods regardless of any assumption or constraint? In this paper, motivated by a two-sample discrepancy problem which uses an estimator with one sample measuring the distribution, the fresh eyes idea is provided: the unlabeled and labeled sets are directly investigated by several similarity measures with the function in the two-sample discrepancy problem theory, leading to a general active learning framework. This framework integrates the informativeness and representativeness into one formula, whose optimization falls into a standard quadratic programming (QP) problem.

To the best of our knowledge, this work is the first attempt to develop a well-defined, complete, and general framework for active learning in the sense that both representativeness and informativeness are taken into consideration based on the two two-sample discrepancy problem. According to the two-sample discrepancy problem, this framework provides a straightforward and meaningful way to measure the representativeness by fully investigating the triple similarities that include the similarities between a query sample and the unlabeled set, between a query sample and the labeled set, and between any two candidate query samples. If a proper function is founded that satisfy the conditions in the two-sample discrepancy problem, the representative sample can be mined with the triple similarities in the active learning process. Since the uncertainty is also the important information in the active learning, an uncertain part is combined the triple similarities with a trade-off to weight the importance between representativeness and informativeness. Therefore, the most significant contribution of our work is that it provides a general idea to design active learning methods, under which various active learning algorithms can be tailored through suitably choosing a similarity measure or/and an uncertainty measure.

Rooted in the proposed framework, a practical active learning algorithm is designed. In this algorithm, the radial basis function (RBF) is adopted to measure the similarity between two samples and then applied to derive the triple similarities. Different from traditional ways, the kernel is calculated by the posterior probabilities of the two samples, which is more adaptive to potentially large variations of data in the whole active learning process. Meanwhile, we modify the Best-vs-Second-Best (BvSB) strategy \cite{JF2012}, which is also based on the posterior probabilities, to measure the uncertainty. We verify our algorithm on fifteen UCI benchmark \cite{AA2010} datasets. The extensive experimental results demonstrate that the proposed active learning algorithm outperforms the state-of-the-art active learning algorithms.

The rest of this paper is organized as follows: Section 2 presents our proposed active learning framework in details. The experimental results as well as analysis are given in Section 3. Section 4 provides further discusses about the proposed method. Finally, we draw our conclusions in Section 5.

\section{The Proposed Framework}
Suppose there is a dataset with ${n}$ samples, initially we randomly divide it into a labeled set and an unlabeled set. We denote that ${{L}_{t}}$ is a training set with ${n_t}$ samples and $ { {U}_{t}}$ is an unlabeled set with ${u_t}$ samples at time ${  {t}}$, respectively. Note that ${{n}}$ is always equal to ${n}_{t}+{u}_{t}$. Let ${ {f}_{t}}$ be the classifier trained on ${{L}_{t}}$. The objective of active learning is to select a sample ${ {x}_{s}}$ from ${{U}_{t}}$ and a new classification model  trained on ${{L}_{t}}$ $\cup$ ${{x}_{s}}$ is learned, which has the maximum generalization capability. Let ${ {Y}}$ be the set of class labels. In the following discussion, the symbols will be used as defined above. We review the basics of the two-sample discrepancy problem below.
\subsection{The two-sample discrepancy problem}
 Define ${ X = \{x_1,..., x_m\}}$, and ${ Z=\{z_1,...,z_n\}}$ are the observations drawn from a domain ${ D}$, and let ${ x}$ and ${ z}$ be the distributions defined on ${ X}$ and ${ Z}$, respectively. The two samples discrepancy problem is to draw i.i.d. sample from ${ x}$ and ${ z}$ respectively to test whether ${ x}$ and ${ z}$ are the same distribution \cite{A2007}. The two-sample test statistic for measuring the discrepancies is proposed by Anderson and Hall  \cite{H1984,AH1994}. In literature \cite{AH1994}, the two-sample test statistics are used for measuring discrepancies between two multivariate probability density functions (pdf). Hall explains the integrated squared error between a kernel-based density of a multivariate pdf and the true pdf. With a brief notation, it can be represented as
\begin{equation}\label{formula21}
{H}=\int(\hat{f}_{\sigma}-f)
\end{equation}
where ${\hat{f}_\sigma}$ denotes the density estimate, ${\sigma}$ denotes the associated bandwith, and ${ f}$ is the true pdf. In such way, a central limited theorem for ${ H}$ which relies the bandwidth within ${\hat{f}_\sigma}$ is derived, and represented in Theorem \ref{th2}.
\newtheorem{theorem}{Theorem}
\begin{theorem}
Define
\begin{displaymath}
G_n(x,y) = E\{Q_n(X_1,x)Q_n(X_1,y)\},
\end{displaymath}
 and assume ${Q_n}$ is symmetric,

 \begin{displaymath}
\begin{split}
&{{Q_n}((X_1,X_2)|X_1)=0}\\
&{E({Q_n^2}(X_1,X_2))<\infty},
\end{split}
\end{displaymath}
for each ${n}$. If
\begin{displaymath}
\lim_{n\rightarrow\infty} \frac{E\{{G_n^2}(X_1,X_2)\}+n^{-1}E\{{Q_n^4}(X_1,X_2)\}}{[E\{{Q_n^2}(X_1,X_2)\}]^2} = 0,
\end{displaymath}
 then
\begin{displaymath}
{U_n\equiv\sum\sum_{1\leq i \leq j\leq n}Q_n(X_i,X_j)}
\end{displaymath}
is asymptotically normally distributed with zero mean and variance given by ${(1/2)n^2E\{{Q_n^2}(X_1,X_2)\}}$.
\label{th2}\\
where ${Q_n}$ is a symmetric function depending on ${n}$ and ${X_1,...,}$ ${X_n}$ are i.i.d. random variables (or vectors). ${U_n}$ is a random variable of a simple one-sample U-statistic.
\end{theorem}

 Following \cite{H1984}, the brief proof is provided in the Appendix A. It is intuitively obvious that ${ H}$ is a spontaneous test statistics for a significance test against the hypothesis that ${ f}$ is indeed the correct pdf. Based on above a theorem, the two-sample versions of ${ H}$ is investigated \cite{AH1994}. Naturally, the objective of the statistic is
\begin{equation}\label{formula22}
{H}_{{\sigma_1}{\sigma_2}}=\int(\hat{f}_{\sigma_1}-\hat{f}_{\sigma_2})
\end{equation}
in which, for ${ j=1,2}$, ${\hat{f}_{\sigma_j}}$ is a density, which is estimated from the ${ j_{th}}$ sample with smoothing parameter ${\sigma_{j}}$. Based on the two-sample versions, which is used to measure two distributions with two samples from them, we develop it to active learning to find a sample to measure the representativeness in labeled dataset and unlabeled dataset, and we call such a problem as ''two-sample discrepancy problem''. The two-sample discrepancy problem is presented in Theorem \ref{th1}, and the proof is provided in the Appendix B.
\begin{theorem}
 Suppose ${\{X_{j1},...,X_{jn_j}\}}$, ${ j = 1,2}$ two independent random samples, from ${ p}$-variate distributions with densities ${f_j}$, ${ j = 1,2}$, and define ${\sigma_{j}}$ is a bandwith and ${ K}$ is a spherically symmetric ${ p}$-variate density function. Assuming
 \begin{displaymath}
  \begin{split}
\lim_{{n_1}\rightarrow\infty}{\sigma_1}=0,\lim_{{n_2}\rightarrow\infty}{\sigma_2}=0,\\
\lim_{{n_1}\rightarrow\infty}{n_1}{\sigma_1}=\infty,\lim_{{n_2}\rightarrow\infty}{n_2}{\sigma_2}=\infty
\end{split}
\end{displaymath}
 and define the estimator of a ${ p}$-variate distribution with density ${ f_j}$ is
\begin{displaymath}
\hat{f_j}= (n_{j}\sigma_j^p)^{-1}\sum_{i=1}^{n_j}K\{(x-X_{ji})/\sigma_j\}
\end{displaymath}
The discrepancy of the two distributions can be measured by two-sample discrepancy problem with a minimum distance ${ n^{-1/2}\sigma^{-p/2}}$, where ${n}$ is ${n_1 + n_2}$.
\label{th1}
\end{theorem}
The other details proofs and theorem about the two-sample discrepancy problem can refer to \cite{AH1994,A2007}. Suppose we find a density function ${F}$ in the theorem 2, following\cite{RZ2012},the empirical estimate of distribution discrepancy between ${X}$ and ${Z}$ with the samples ${x_i \in X}$ and the samples ${z_i \in Z}$ can be defined as follows:

\begin{equation}\label{formulate23}
\begin{split}
F(x_i)-F(z_i)=sup\Arrowvert(m_{j}\sigma_j^p)^{-1}\sum_{i=1}^{m_j}F\{(x-x_{i})/\sigma_j\}\\
-(n_{j}\sigma_j^p)^{-1}\sum_{i=1}^{n_j}F\{(z-z_{i})/\sigma_j\}\Arrowvert
\end{split}
\end{equation}

In Eq.(3), it shows that given a sample $x \in X$ and a sample ${z \in Z}$, the distribution discrepancy of the two data set can be measured with the two sample under the density function $F$. According to theorem 2, Eq.(3) has a minimum distance between the two data sets. If the upper bound can be minimized, the distribution of X and Z will have the similar distribution with the density function $F$, where $F$ is rich enough{\cite{RZ2012}}.
 According the review above, the consistency of distribution between two sets can be measured similarity with a proper density function $F$. If we treat ${X}$ as the unlabeled set and ${Z}$ as the labeled set, we can discover that the two-sample discrepancy problem can be adapted to the active learning problem. If a sample in the unlabeled set can measure the distribution of the unlabeled set, adding it to the labeled set, and removing it from the unlabeled set without changing the distribution of the unlabeled set, it can make the two sets have the same distribution. In active learning, there only a small proportion of the samples are labeled, so the finite sample properties of the distribution discrepancy are important for the two-sample distribution discrepancy to apply to active learning. We briefly introduce one test for the two-sample discrepancy problem that has exact performance guarantees at finite sample sizes, based on uniform convergence bounds, following the literature \cite{A2007}, and the McDiarmid \cite{M1989} bound on the biased distribution discrepancy statistic is used. With the finite sample setting, we can establish two properties of the distribution discrepancy, from which we derive a hypothesis test. Intuitively, we can observe that the expression of the distribution discrepancy and MMD are very similar. If we define $\sigma$= 1, they have the same expression. First, if we let $m_j \sigma^p=m$ and $n_j \sigma_j^p=n$, according to\cite{A2007}, it is shown that regardless of whether or not the distributions of two data sets are the same, the empirical distribution discrepancy converges in probability at rate $O((m_j \sigma^p+n_j \sigma_j^p)^{(-1/2)})$, which is a function of $\sigma$. It shows the consistency of statistical tests. Second, probabilistic bounds for large deviations of the empirical distribution discrepancy can be given when the two data sets have the same distribution. According to theorem 2 in our paper, these bounds lead directly to a threshold. And according to the theorems in section 4.1 of the convergence rate is perspective and the biased bound links with $\sigma$. The more details about the properties with finite samples can refer to \cite{A2007}. Hence, this theorem. 2 can be used to measure the representiveness of an unlabeled sample in active learning.

\subsection{The General Active Learning Framework}
In our proposed framework, the goal is to select an optimal sample that not only furnish useful information for the classifier ${{f}_{t}}$, but also shows representativeness in the unlabeled set ${ {U}_{t}}$ , and as little redundancy as possible in the labeled set ${{L}_{t}}$. In such a way, the sample ${{x}_{s}}$ should be informative and representative in the unlabeled set and labeled set, i.e., if we query the sample without representativeness the two samples  queried from two iterations separately furnish useful information but they may provide same information, then one of them will become redundant. Hence, our proposed framework aims at selecting the samples with different pieces of information. To achieve this goal, an optimal active learning framework is proposed, which combines the informativeness and representativeness together,with a tradeoff parameter that is used to balance the two criteria.

For the representative part, the two-sample discrepancy problem is used. As reviewed above, The two-sample discrepancy problem is used to examine $H_{12}$ in Eq.(2) under the hypothesis $\hat{f}^{\sigma1} = \hat{f}^{\sigma2}$, and the objective is to minimize $H_{12}$. So the essential problem of the distribution discrepancy is to estimate the $\hat{f}^{\sigma1}$ and $\hat{f}^{\sigma2}$. In theorem 2, it shows a way to find an estimator of a p-variate distribution with $f_j$, which is estimated from the $j_{th}$ sample, so a sample ${x_j}$ in a data set can always obtain an estimator of a p-variate distribution with $f_j$. If we use the $x_j$ to find two estimators of a p-variate distribution with two data sets A and B, the distribution discrepancy of A and B can be measured by the difference of the two estimators. Let A and B represent the labeled data and the unlabeled data respectively in active learning, and a sample in unlabeled data can obtain two estimators with the labeled data and the unlabeled data, respectively. For each unlabeled sample, a distribution discrepancy can be obtained between the labeled data and the unlabeled data. Distribution of labeled data and that of unlabeled data are respectively corresponding to the term $M_2$ and $M_3$ in the formula (4). If the difference between $M_2$ and $M_3$ is small, it indicates that when the sample is added to the labeled data, it will decrease the distribution discrepancy of the unlabeled data and labeled data. For representativeness, our goal is also to find the sample that makes the distribution discrepancy of unlabeled data and labeled data small. However, exhaustive search is not feasible due to the exponential nature of the search space. Hence, we solve this problem using numerical optimization-based techniques. We define a binary vector ${\alpha}$ with ${u_t}$ entries, and each entry ${\alpha_i}$ is corresponding to the unlabeled point ${x_i}$. If the point is queried, the ${\alpha_i}$ is equal to 1, else 0. The discrepancy of the samples in the unlabeled set is also a vector with ${u_t}$ entries. For each sample ${x_i}$ in the unlabeled , we measure the discrepancy with two parts, which are defined ${M_2(i)}$ and ${M_3(i)}$ as the distribution in labeled set and unlabeled set respectively. The distribution discrepancy of sample ${x_i}$ can be measured as
\begin{displaymath}
M_2(i)-M_3(i)
\end{displaymath}
 Meanwhile, we want to make sure the sample is optimal in the latent representative samples. And a similarity matrix ${M_1}$ is defined with ${u_t \times u_t}$ entries whose ${(i,j)^{th}}$ entry represents the similarity between ${x_i}$ and ${x_j}$ in unlabeled set. Hence, the optimization problem of representative part can be formulated as follows:
\begin{equation}\label{formula24}
\min_{\alpha^T{1^{u_t}}=1,\alpha_i\in\{0,1\}}\alpha^T{M_1}\alpha+\alpha^T({M_2-M_3})
\end{equation}
For the informative part, we compute an uncertainty vector ${C}$ with ${u_t}$ entries, where ${C(x_i)}$ denotes the uncertainty value of point ${x_i}$ in the unlabeled set. Therefore, combining the representative part and uncertain part with a tradeoff parameter, we can directly describe the objective of active learning framework and formulate as follows:
\begin{equation}\label{formula1}
\begin{split}
    \min_{\alpha}\alpha^{T}M_{1}\alpha+{\alpha}^{T}(M_{2}-M_{3})+{\beta}C\\
    {s.t.}~~{\alpha}^{T}1^{{u}_{t}}=1,{\alpha}_{i} \in {\{0,1\}}
    \end{split}
\end{equation}

If ${S}$ is a spherically symmetric density function to measure the similarity between two samples, the entry in ${ {M}_{1}, {M}_{2}, {M}_{3}}$ can be defined as below. In the proposed framework, ${M_1}$ is a matrix with ${R}^{{u}_{t}\times{u}_{t}}$, and the entry in it can be formulated as follows:
\begin{equation}\label{formula2}
    {M}_{1}(i,j)=\frac{1}{2}S(i,j)
\end{equation}
${M_1(i,j)}$ is the similarity between the ${i}^{th}$ and ${j}^{th}$ sample in unlabeled set. However, differing from the entry in ${M_1}$, the entry in ${ {M}_{2}}$ measures the distribution between one sample and the labeled set. The formulation can be written as:
\begin{equation}\label{formula3}
    {M}_{2}(i)=\frac{{n}_{t}+1}{n}\sum_{j=1}^{{n}_{t}}S(i,j)
\end{equation}
 where $({n}_{t}+1)/{n}$ is a weight, corresponding to the percentage of the labeled set in the whole data set, which is used to balance the importance of the unlabeled data and labeled data. It represents the similarity between sample ${x}_{i}$ in the unlabeled set and the labeled set. If ${M}_{2}(i)$ in ${M}_{2}$ is smaller, it implies that the sample ${x}_{i}$ chosen from the unlabeled set is more different with the labeled samples, and the redundancy is also reduced. Therefore, ${M}_{2}$ is to ensure the selected samples contain more diversity.
Similar to the definition of ${M}_{2}$, we define ${M}_{3}\in {R}^{{u}_{t}\times1}$ to measure the distribution between the sample and the unlabeled set as follows:
\begin{equation}\label{formula4}
    {M}_{3}(i)=\frac{{u}_{t}-1}{n}\sum_{j=1}^{{u}_{t}}S(i,j)
\end{equation}
where $({u}_{t}-1)/{n}$ is a weight corresponding to the percentage of the unlabeled set in the whole data set which has the same meaning with the weight of ${M}_{2}$. It enforces the query sample to present certain similarity to the remaining ones in ${U}_{t}$.

As the description above, ${M}_{1}, {M}_{2}, {M}_{3}$ together can help to select a sample that is representative in unlabeled set and presents low redundancy in labeled set, and this may be an excellent way to combine them together to announce the representativeness of a sample.
In order to ensure the selected sample is also highly informative, ${C}$ is computed as an uncertain vector ${C}\in R^{{u}_{t}\times1}$  of length ${u}_{t}$. Each entry ${C}(i)$ denotes the uncertainty of ${x}_{i}$ in the unlabeled set for the current classifier ${f}^{t}$. It can be formulated as:
\begin{equation}\label{formula5}
    {C}(i)=\ell({f}^{t},{x}_{i}),{x}_{i}\in {U}_{t}
\end{equation}
where $\ell(.)$ is a function to measure the uncertainty of ${x}_{i}$ based on ${f}^{t}$. Based on the fundamental idea in the proposed framework, if we can find a reasonable function to measure the similarity between two samples and a method to compute the uncertainty of a sample, an active learning algorithm can be designed reasonably. Meanwhile, we can find that the formulation of our proposed framework is an integer programming problem which is NP hard due to the constraint ${{\alpha}_{i}\in {\{0,1\}}}$ . A common strategy is to relax ${\alpha}_{i}$ to a continuous value range [0, 1], and we will derive a convex optimization problem as:
\begin{equation}\label{formula6}
\begin{split}
    \min_{\alpha}\alpha^{T}M_{1}\alpha+{\alpha}^{T}(M_{2}-M_{3})+{\beta}C\\
    {s.t.}~~{\alpha}^{T}1^{{u}_{t}}=1,{\alpha}_{i} \in {[0,1]}
    \end{split}
\end{equation}

This is a quadratic programming (QP) problem, and it can be solved efficiently by a QP solver. Once we solve ${\alpha}$   in formula (\ref {formula6}), we set the largest element ${\alpha}_{i}$  to 1 and the remaining ones to 0. Therefore, the continuous solution can be greedily recovered to the integer solution.
\subsection{The Proposed Active Learning Method}
Based on the proposed optimal framework, we propose an active learning method relying on the probability estimates of class membership for all the samples.
\subsubsection{Computing Representative Part}
 The proposed framework is a convex optimization problem, which requires the similarity matrix to be positive semi-definite. Kernel matrix has been has been widely used as the similarity matrix, which maintains the convexity of the objective function by constraining the kernel to be positive semi-definite \cite{SR2014,RZ2012,PR2009}. Without losing generality, we adopt the Radial Basis Function to measure the similarity between two samples. Generally speaking, the similarity matrices are usually directly computed based on the Euclidean distance of two samples with RBF in feature space \cite{SR2014,RZ2012,LZ2015}. However, in practice, the distribution of the samples in unlabeled set and that of samples in labeled set may be different as the number of the two sets is changing. Therefore, it may not be reasonable to use such kernel matrix to measure the similarity in the active learning process. Hence, in our proposed method, we measure the similarity between two samples using the posterior possibility combined with RBF. The posterior probability represents the importance of a sample with respect to a classifier. If a similarity is measured based on the probability kernel, it represents whether the samples have the same impact on the respective classifiers, which can help directly to construct proper classifiers. Thus, the representative samples we select with the probability kernel are effective to enhance the classifiers. Meanwhile, the probability leads to a distribution of samples, which is denser than that of feature similarity. Therefore, the redundancy can be reduced and more informative samples can be queried then. To compute the posterior probability, the classifier ${f}^{t}$ is applied on the sample ${x}_{i}$ in the unlabeled set to yield the posterior probability ${p}^{i}_{c}$ with respect to the corresponding class ${c}$, where ${c}$ belongs to the labels set ${Y}$ [11]. Let ${P}^{i}$ and ${P}^{j}$ be the posterior probability of two samples in the unlabeled set, ${P}^{i} = \{{p}^{i}_{1},{p}^{i}_{2},...,{p}^{i}_{|Y|}\}$, where $|Y|$ is the number of classes. Then, the ${(i,j)}^{th}$ entry in ${M}_{1}$ can be represented as:
\begin{equation}\label{formula7}
   \begin{split}
     {M}_{1}(i,j)=\frac{1}{2}S(i,j)=\frac{1}{2}exp(-\gamma\ast\|{P}^{i}-{P}^{j}\|^{2}_{2})\\
     s.t. \sum_{k=1}^{|Y|}{p}_{k}^{i}=1,\sum_{k=1}^{|Y|}{p}_{k}^{j}=1
    \end{split}
\end{equation}

where $\gamma$ is the kernel parameter. Thus, the element in ${M}_{2}$ can be computed as:
\begin{equation}\label{formula8}
    {M}_{2}(i)=\frac{{n}_{t}+1}{n}\sum_{j=1}^{{n}_{t}}exp(-\gamma\ast\|{P}^{i}-{P}^{j}\|^{2}_{2})
\end{equation}

Meanwhile, the ${i}^{th}$ entry in ${M}_{3}$ can be figured out as:
\begin{equation}\label{formula9}
    {M}_{3}(i)=\frac{{u}_{t}-1}{n}\sum_{j=1}^{{u}_{t}}exp(-\gamma\ast\|{P}^{i}-{P}^{j}\|^{2}_{2})
\end{equation}
\subsubsection{Computing Uncertainty Part}

\begin{figure*}\label{fig3}
\begin{center}
\epsfig{file = 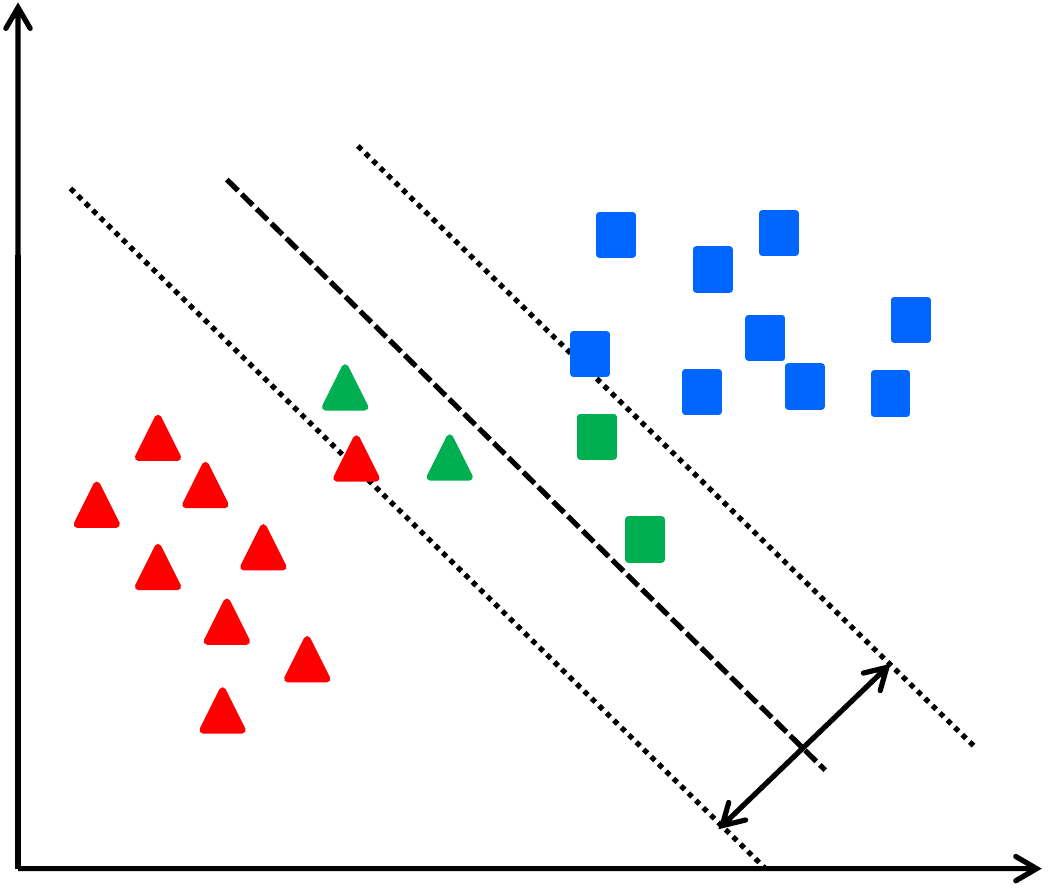, height = 1.8 in, keepaspectratio}
\epsfig{file = 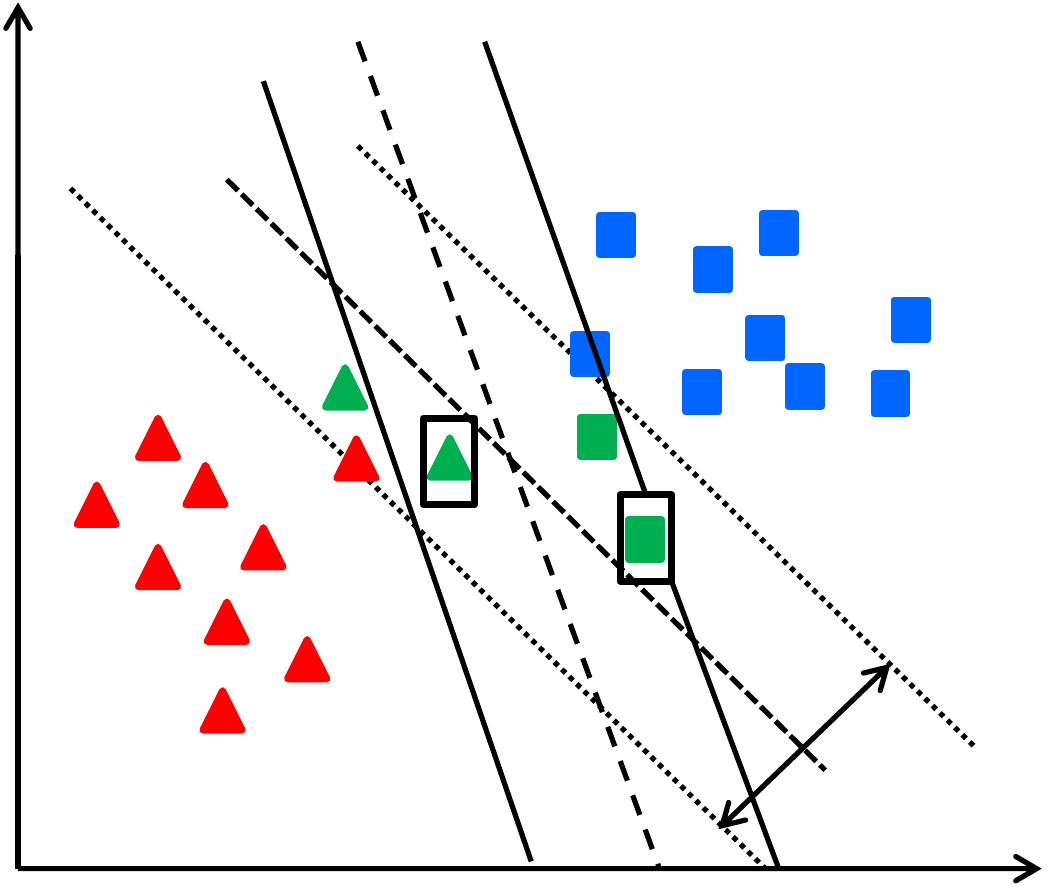, height = 1.8 in, keepaspectratio}
\epsfig{file = 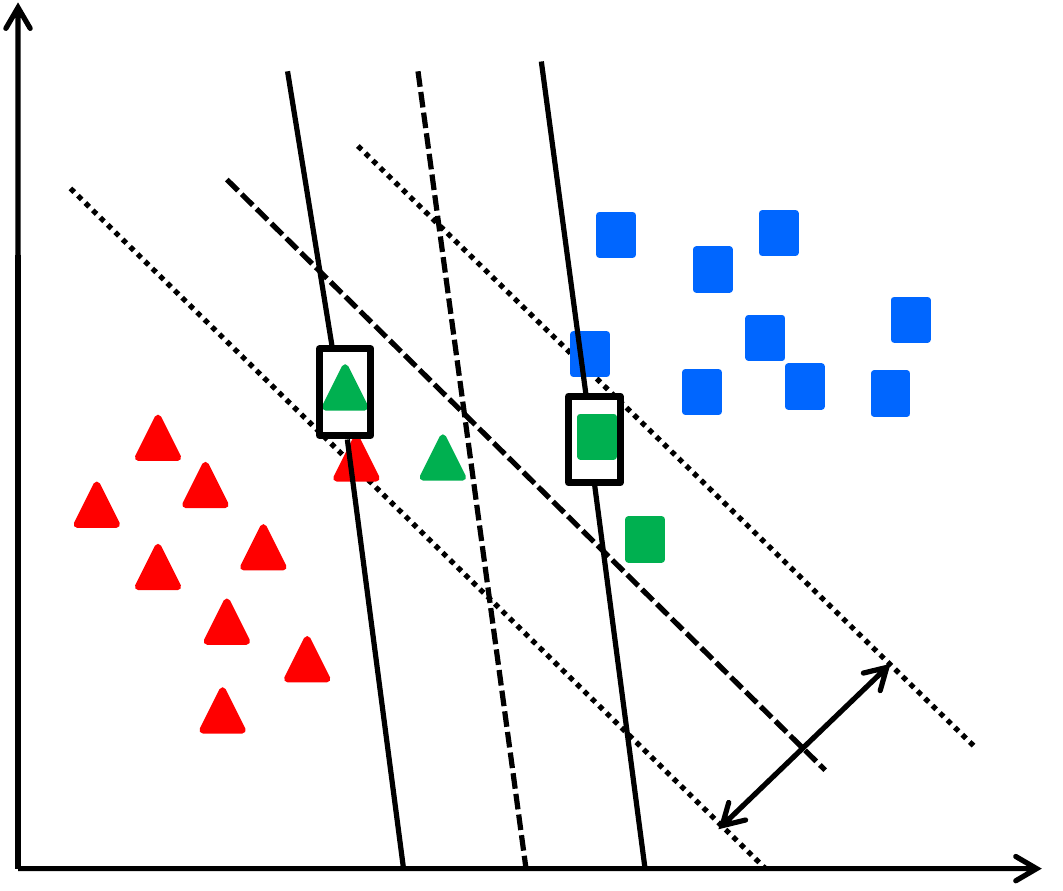, height = 1.8 in, keepaspectratio}\\
\caption{The triangle and the rectangle are two classes, the green points are the unlabeled set. The first one is the original model, the green points with black rectangle in median one are the samples queried with BvSB, the green points with black rectangle in last one are the samples queried with the proposed uncertain method.}
\end{center}
\end{figure*}

The uncertain part is used to measure the informativeness of the sample for the current classifier ${f}^{t}$. If it is hard for  ${f}^{t}$ to decide a sample' s class membership, it suggests that the sample contains a high uncertainty. So it may probably be the one that we want to select and label. In this part, we design a new strategy to measure the uncertainty of a sample, which is a modification of the BvSB strategy. The BvSB is a method based on the posterior probability, which considers the difference between the probability values of the two classes with the highest estimated probability and has been described in details in \cite{AF2009}. Such a measure is a direct way to estimate the confusion about class membership from the classification perspective. The mathematical form is following: for a sample ${x}_{i}$ in the unlabeled dataset, let ${p}_{h}^{i}$ be the maximum value in ${P}^{i}$ for class ${h}$, also let ${p}_{g}^{i}$ be the second maximum value in ${P}^{i}$ for class ${g}$. The BvSB measure can be obtained as follow:
\begin{equation}\label{formula10}
    {d}_{BvSB}^{i}={p}_{h}^{i}-{p}_{g}^{i}
\end{equation}

The smaller BvSB measure is the higher uncertainty of the sample. But such a method is inclined to select samples close to the separating hyperplane. In our proposed method, the SVM classifier is used, hence, we also hope the distance between two classes is large. Therefore, the query sample should not only be close to the hyperplane, but also close to the support vectors. We define such information as position measure. Based on such an idea, we modify the BvSB method to reveal the highly uncertain information behind the unlabeled samples. We denote such the support vectors set as ${SV} = {\{{SV}_{1}, {SV}_{2},..., {SV}_{m}\}}$, where ${m}$ is the number of support vectors. Suppose ${P} = {\{{P}_{1}, {P}_{2},..., {P}_{m}\}}$ is the probability set of support vectors. For each sample ${x}_{i}$ in unlabeled dataset, we construct a similarity function to calculate the distance between the sample and the support vectors with the estimated probability as the position measure:
\begin{equation}\label{formula11}
    {f}({x}_{i},{SV}_{j})=exp(\|{P}^{i}-{P}_{j}\|_{2}^{2})
\end{equation}

If the sample is close to the support vector,${f}({x}_{i},{SV}_{j})$ will be also small. We choose the smallest value of (\ref {formula11}) between the closest support vector and the sample as the position measure of the sample in the classification interval of SVM. The closest support vector can be found as follows:
\begin{equation}\label{formula12}
    {SV}_{c}=\argmin_{{SV}_{j}\in{SV}}({f}({x}_{i},{SV}_{j}))
\end{equation}

\begin{algorithm}[H]
\caption{The proposed algorithm based on the general active learning framework}
\label{alg:Framwork}
\begin{algorithmic}[1]
\Require
 ${L_t =\{(x_i,y_i)\}}$ with ${n_t}$ labeled samples, ${U_t = \{x_i\}}$ with ${u_t}$ unlabeled samples,
   t=0, the trade-off parameter ${\beta}$, the terminating condition ${\delta}$
\Repeat
\State calculate the estimated probability for the samples in ${{L_t}\bigcup{U_t}}$ using LIBSVM.
\State acquire the ${M_1, M_2}$, and ${M_3}$ with the estimated probability according function (\ref{formula7}), (\ref{formula8}) and (\ref{formula9}).
\State calculate the BvSB value for each sample in ${U_t}$ with function (\ref{formula10}).
\State find the closet support vector of each sample in ${U_t}$ with function (\ref{formula11}) and (\ref{formula12}), then, calculate the position measure with the closest support vector.
\State combine the uncertain value of each sample in ${U_t}$ with function (\ref{formula13}).
\State optimize the objective function (\ref{formula6}) ${w.r.t}$ ${\alpha}$ using QP; set the largest elements in ${\alpha}$ to 1 and the others to 0, set the query sample as ${\ x_s}$ with an oracle label.
\State update labeled set ${L_t}$ and unlabeled set ${U_t}$
\Until {The terminating condition ${\delta}$ is satisfied}
\end{algorithmic}
\end{algorithm}

Since our goal is to enhance the classification hyperplanes as well as to improve the classification interval of SVM classifier, we combine the BvSB measure and the position measure together as the uncertainty:
\begin{equation}\label{formula13}
    C({x}_{i})= {d}_{BvSB}^{i}\ast{f}({x}_{i},{SV}_{c})
\end{equation}

By minimizing ${C}({x}_{i})$, the uncertain information is enhanced compared to BvSB. Fig.1 shows the data point selected by the BvSB and the proposed uncertain method. It is worth noting that in the proposed method the main task we need to do is to calculate the estimated probabilities, with which the active learning procedure can be easily implemented. We use the LIBSVM toolbox \cite{CC2001} for classification and probability estimation for the proposed method. Hence, the implementation of our proposed method is simple and efficient.The key steps of the proposed algorithm are summarized in Algorithm 1.

From the descriptions in Section 2.1 and Section 2.2, it can be found that our proposed framework can be generalized to different AL algorithms.

\section{Experiments}

In our experiments, we compare our method with random selection and state-of-the-art active learning methods. All the compared methods in the experiments are listed as follows:
\begin{table}[hb]\label{table1}
\footnotesize
\centering
\caption{Characteristics of the datasets, including the numbers of the corresponding features and samples.}
\begin{tabular}{|c|c|c|} \hline
Dataset& Feature& Instance\\ \hline
australian&14&690\\ \hline
sonar&60&208\\ \hline
diabetis&8&768\\ \hline
german&20&1000\\ \hline
heart&13&270\\ \hline
splice&60&2991\\ \hline
image&18&2086\\ \hline
iris&5&150\\ \hline
monk1&6&432\\ \hline
vote&16&435\\ \hline
wine&13&178\\ \hline
ionosphere&34&351\\ \hline
twonorm&20&7400\\ \hline
waveform&21&5000\\ \hline
ringnorm&20&7400\\ \hline
\end{tabular}
\end{table}

\begin{table*}\label{table2}
\footnotesize
\centering
\caption{Win/Tie/Loss counts of our method versus the competitors based on paired t-test at 95 percent significance level.}
\begin{tabular}{|c|c|c|c|c|c|} \hline
Dataset&Vs BMDR&Vs QUIRE&Vs MP&MARGIN&Vs RANDOM\\ \hline
australian&36/64/0&76/24/0&81/19/0&86/14/0&85/15/0\\ \hline
sonar&54/46/0&80/20/0&72/23/5&73/26/1&77/21/2\\ \hline
diabetis&74/21/5&95/5/0&97/3/0&97/3/0&97/3/0\\ \hline
german&31/67/2&61/39/0&74/26/0&85/15/0&81/17/2\\ \hline
heart&40/58/2&54/44/2&73/27/0&67/31/2&64/36/0\\ \hline
splice&73/24/3&100/0/0&100/0/0&100/0/0&100/0/0\\ \hline
image&60/38/2&81/15/4&100/0/0&98/2/0&95/5/0\\ \hline
iris&64/33/3&77/23/0&67/30/3&22/73/5&37/60/3\\ \hline
monk1&74/26/0&93/7/0&100/0/0&100/0/0&99/1/0\\ \hline
vote&50/48/2&84/14/3&75/25/0&55/42/3&80/19/1\\ \hline
wine&28/70/2&41/55/4&49/49/2&33/67/0&72/28/0\\ \hline
ionosphere&55/42/3&91/8/1&81/17/2&89/8/3&75/24/1\\ \hline
twonorm&0/96/4&94/6/0&97/3/0&85/15/0&100/0/0\\ \hline
waveform&36/44/20&100/0/0&100/0/0&100/0/0&100/0/0\\ \hline
ringnorm&83/17/0&100/0/0&100/0/0&100/0/0&100/0/0\\ \hline
\end{tabular}
\end{table*}

\begin{enumerate}
\item RANDOM: randomly select the selected samples in the whole process.
\item BMDR: Batch-Mode active learning  by querying Discriminative and Representative samples, active learning to select discriminative and representative samples by adopting maximum mean discrepancy to measure the distribution difference and deriving an empirical upper bound for active learning risk \cite{ZJ2013}.
\item QUIRE: min-max based active learning, a method that queries both informative and representative samples \cite{SR2014}.
\item MP: marginal probability distribution matching based active learning, a method that prefers representative samples \cite{RZ2012}.
\item MARGIN:simple Margin, active learning that selects uncertain samples that based on the distance the point to the hyperplane \cite{SD2001}.
\end{enumerate}

Note that the BMDR and MP are batch-mode active learning method in the original literature \cite{RZ2012,ZJ2013}, so we set the batch size as 1 to select a single sample to label at each iteration as in \cite{SR2014}. Following the previous active learning publications \cite{SR2014,SD2001,RZ2012,ZJ2013} we verify our proposed method on fifteen UCI benchmark datasets: australian, sonar, diabetis, german, heart, splice, image, iris, monk1, vote, wine, ionosphere, twonorm, waveform, ringnorm, and their characteristics are summarized in Table 1.

In our experiments, we divide each dataset into two parts as a partition 60\% and 40\% randomly. We treat the 60\% data as the training set and 40\% data as the testing set \cite{ZJ2013}. The training set is used for active learning and the testing data is to compare the prediction accuracy of different methods. We can start our proposed method without labeled samples, but for MARGIN method, the initial labeled data is obligatory. Hence, insuring a fair comparison for each method, we start all the active learning methods with an initially labeled small dataset which is just enough to train a classifier. In our experiments, for each dataset, we select just one sample from each class as the initially labeled data. Same to \cite{ZJ2013}, we select them from the training dataset. The rest of the training set is used as the unlabeled dataset for active learning. For each dataset, the procedure is stopped when the prediction accuracy does not increase for any methods, or the proposed method keeps outperforming the compared methods after several iterations. This stopping criterion ensures that the proposed method and the compared methods have a good contrast and also decrease the labeling cost. As to the compared methods' parameters setting, we use the values in the original papers.  In the proposed method, there is a trade-off parameter $\beta$. Following the previous work \cite{SR2014,ZJ2013}, we choose the best value of $\beta$ from a candidate set by cross validation at each iteration. For all methods, the SVM is used as the classifier, and the LIBSVM toolbox \cite{CC2001} is used. We choose the RBF kernel for the classifier, and the same kernel width is used for the proposed algorithm and the comparison methods.The parameters of SVM are adopted with the empirical values\cite{CC2001}. Since our proposed method is a QP problem, we can solve it with QP toolbox. In our experiments, we use the MOSEK toolbox\footnote{https://mosek.com/} to solve our optimization problem.

We conduct our experiments in 10 runs for each dataset on each active learning method, and show the average performance of each method in Figure. 2. In active learning field, the performances of the entire query process of the competing methods are usually presented for comparison. Besides, we compare the competing methods in each run with the proposed method based on the paired t-test at 95 percent significance level \cite{SR2014,ZJ2013}, and show the Win/Tie/Lose for all datasets in TABLE 2.

\begin{figure*}\label{fig1}
\begin{center}
\epsfig{file = 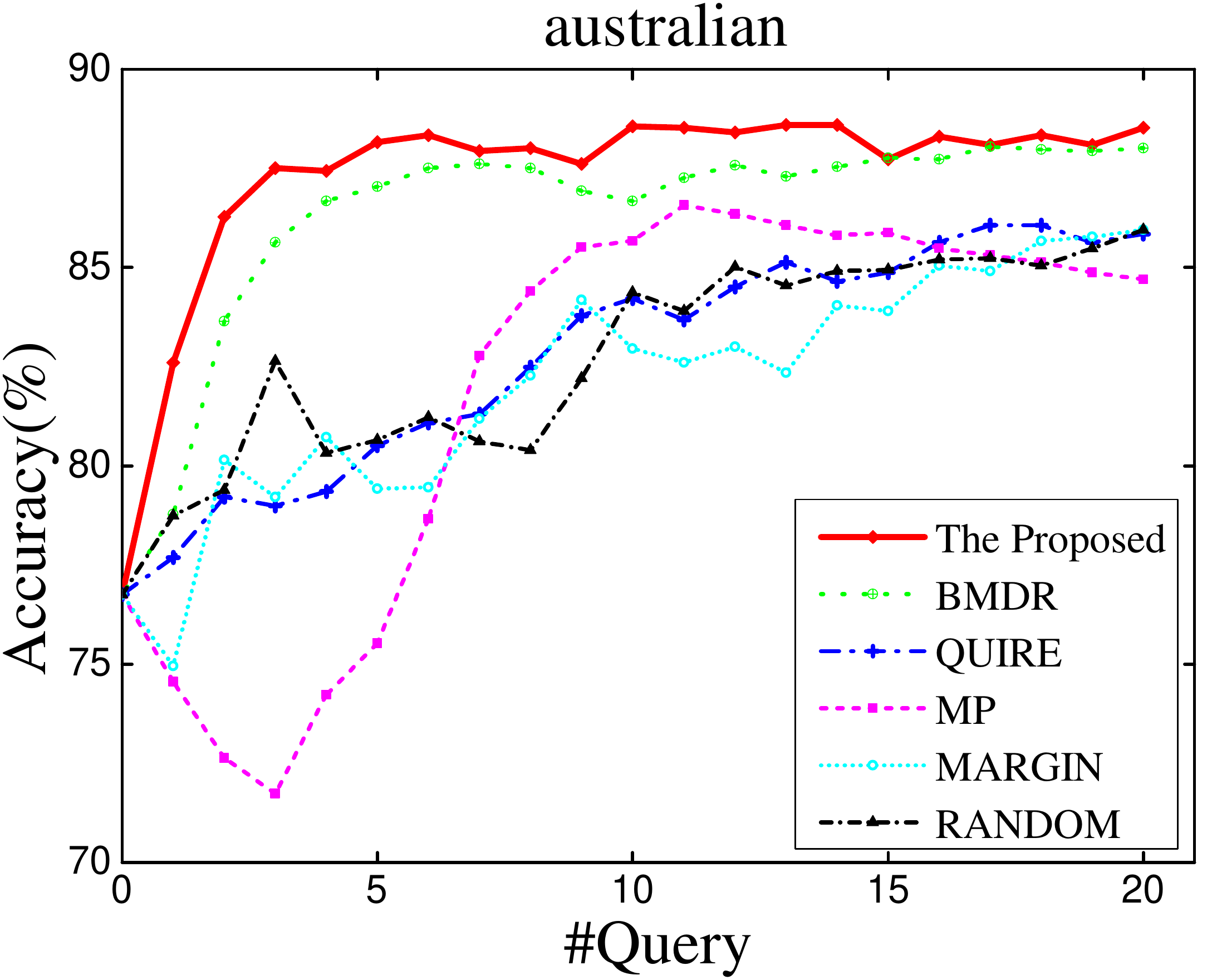, height = 1.8 in, keepaspectratio}
\epsfig{file = 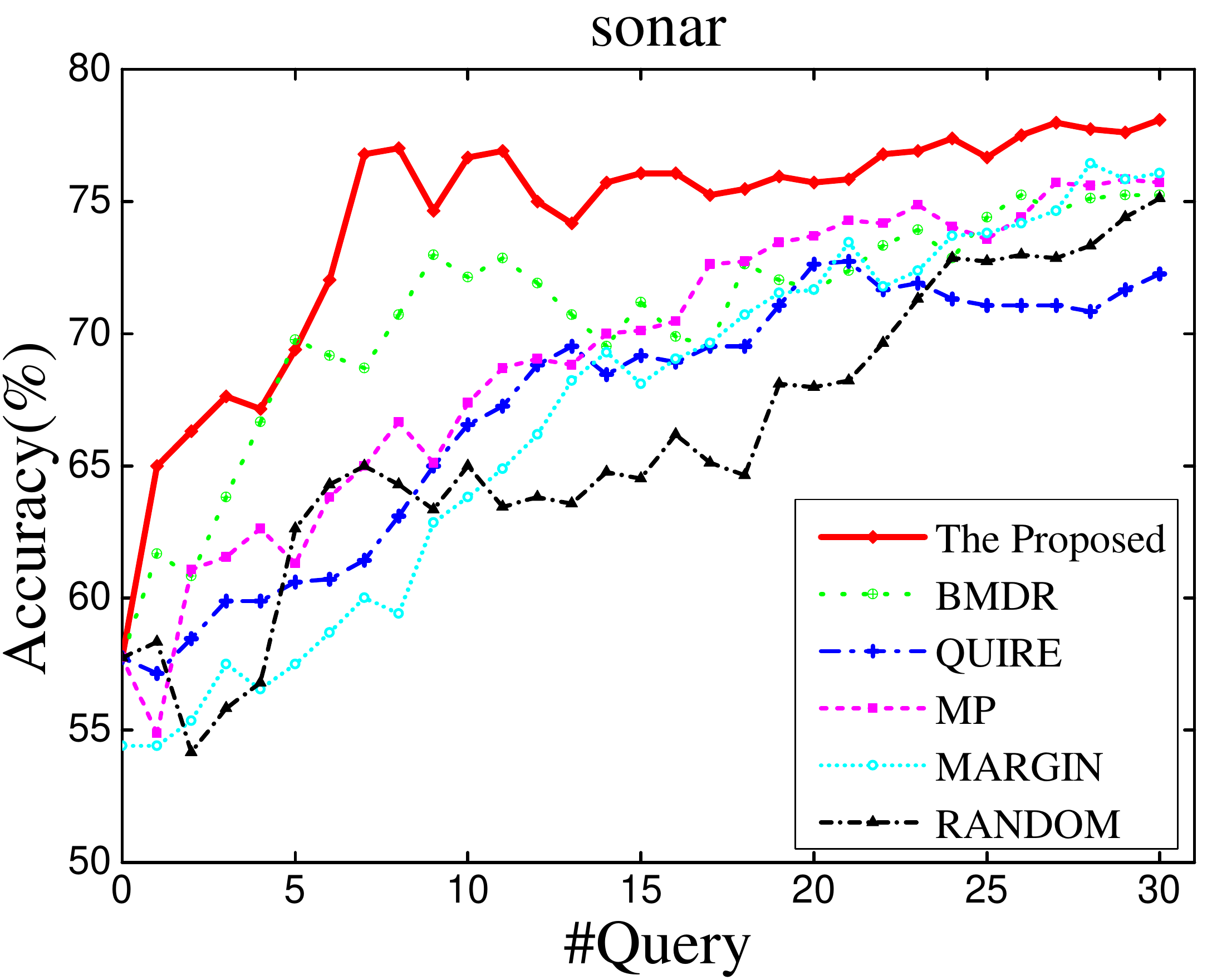, height = 1.8 in, keepaspectratio}
\epsfig{file = 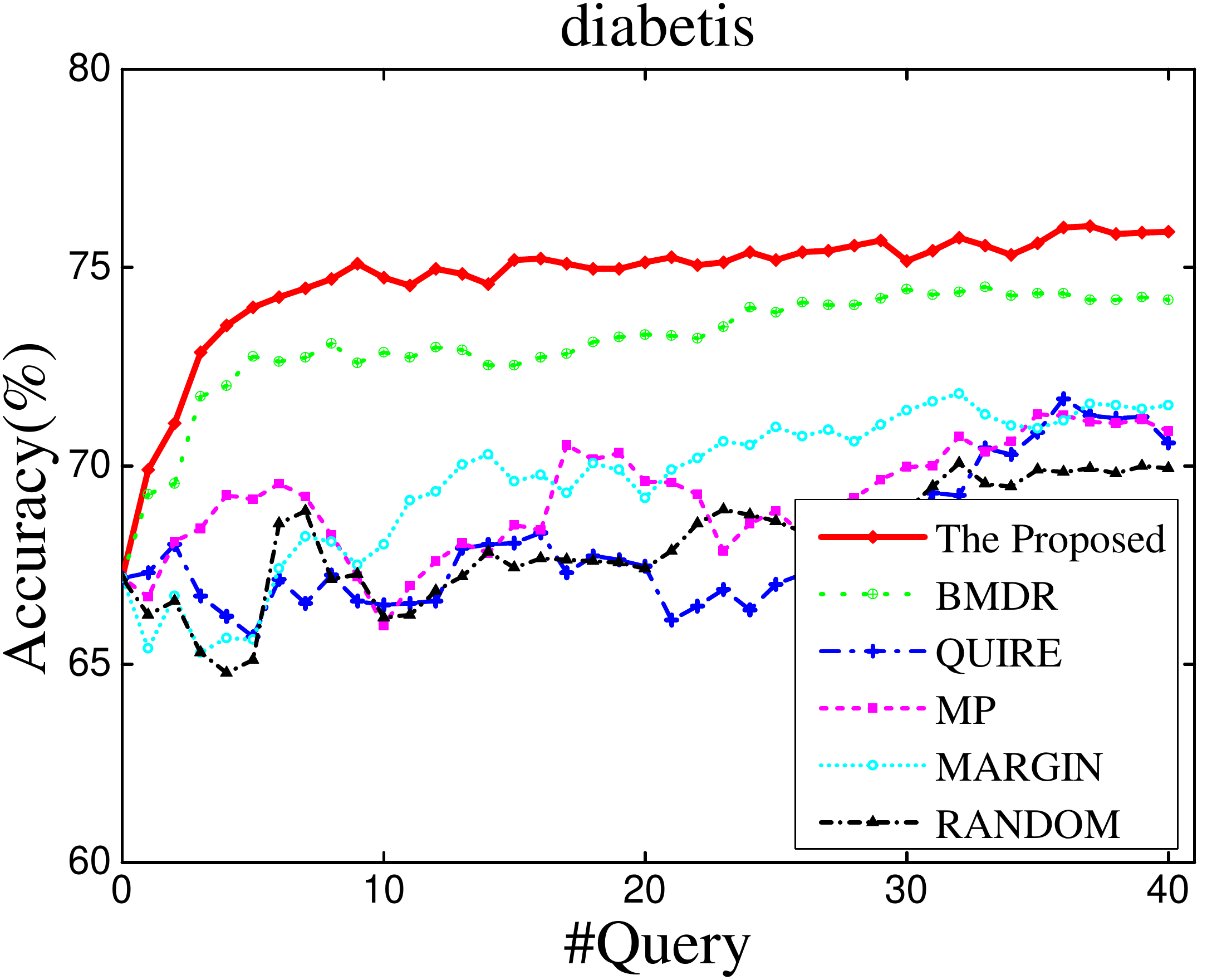, height = 1.8 in, keepaspectratio}\\
\epsfig{file = 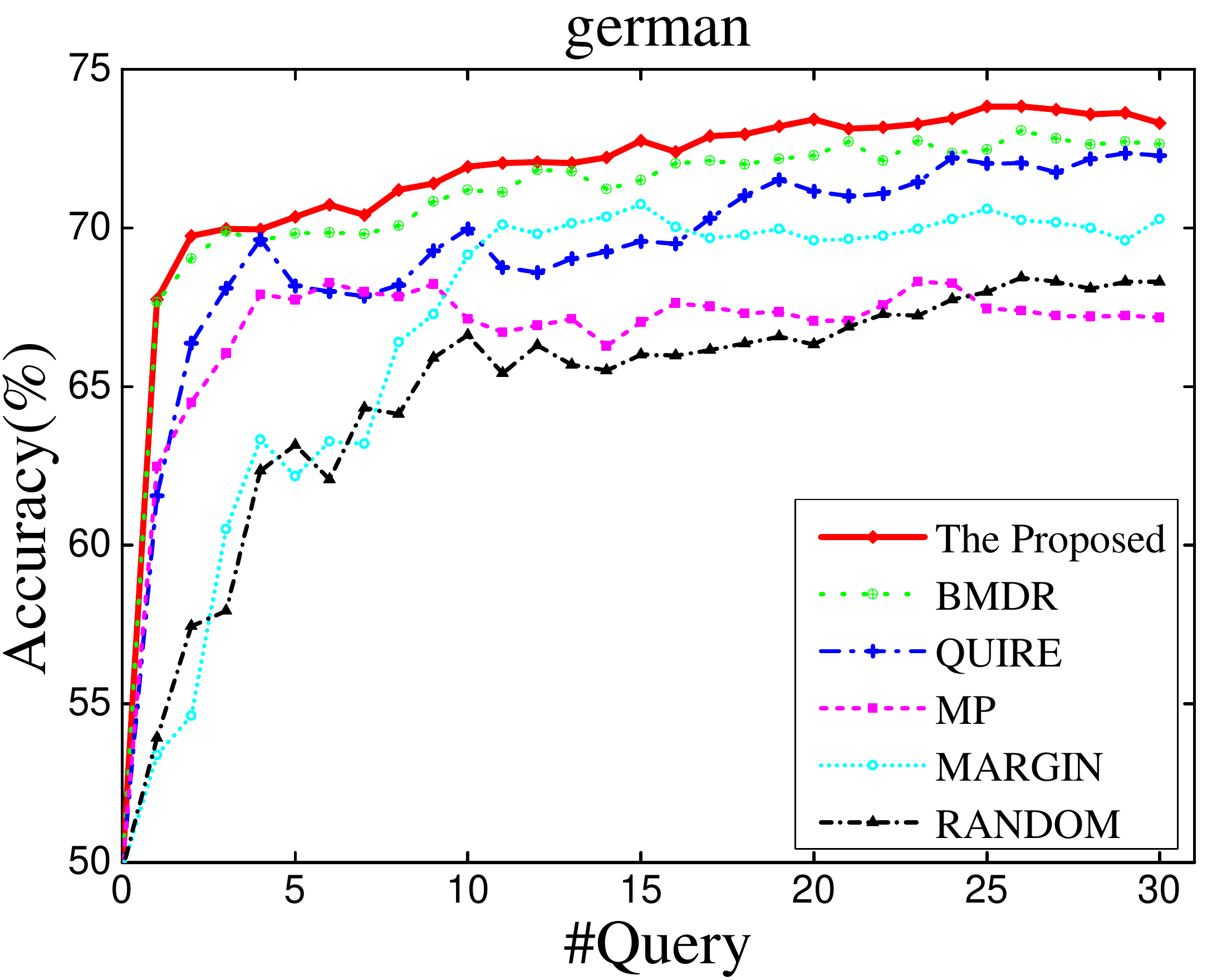, height = 1.8 in, keepaspectratio}
\epsfig{file = 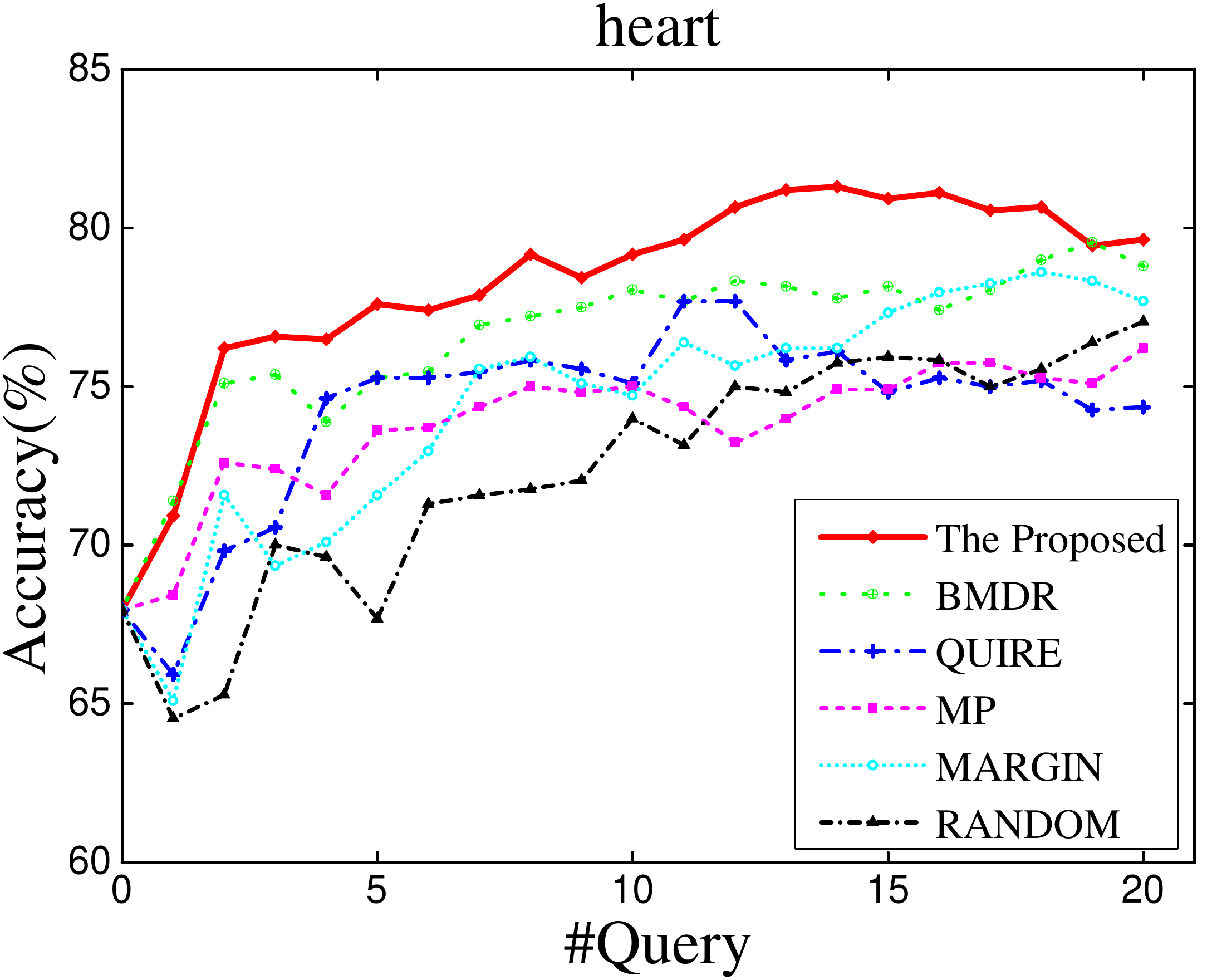, height = 1.8 in, keepaspectratio }
\epsfig{file = 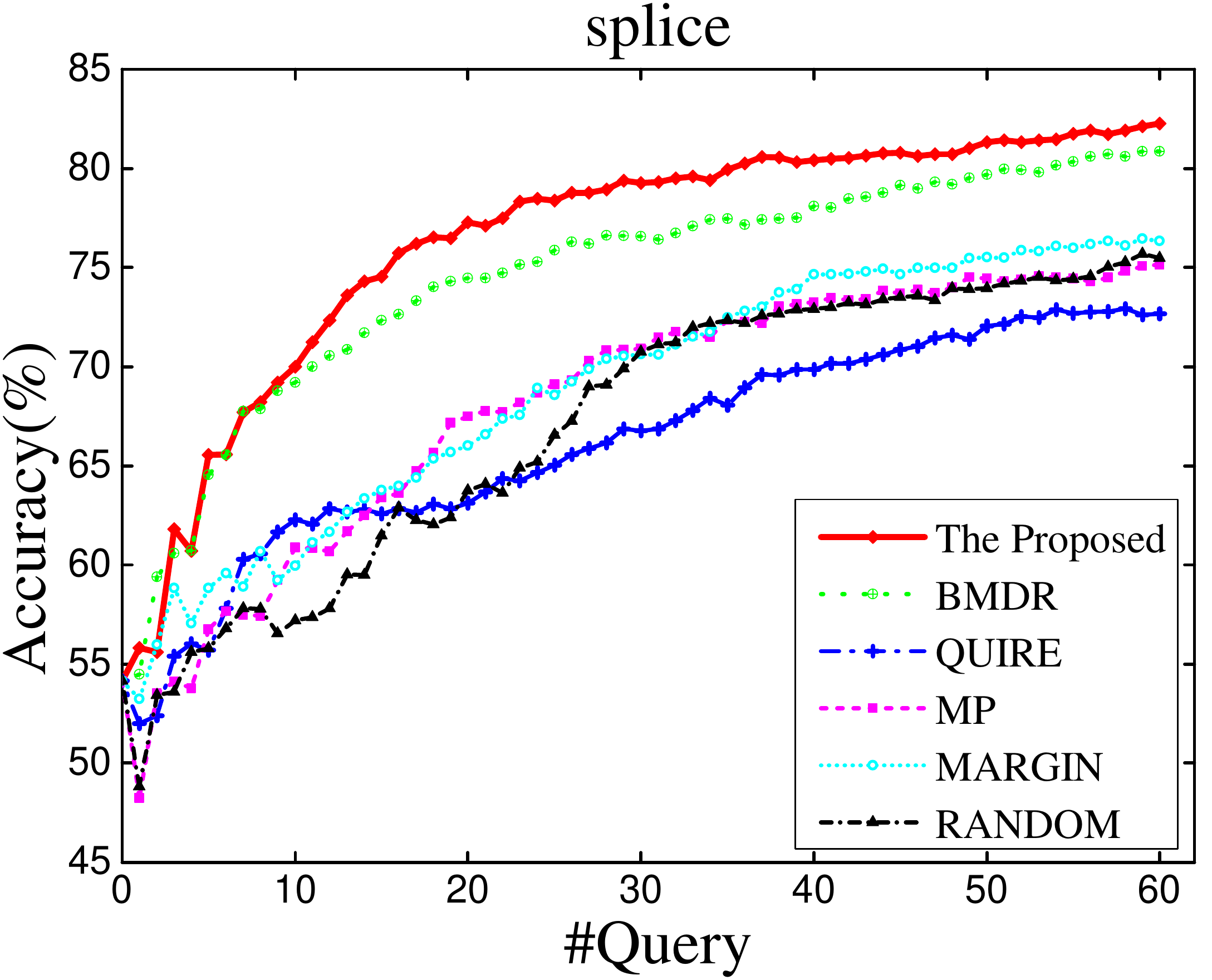, height = 1.8 in, keepaspectratio }\\
\epsfig{file = 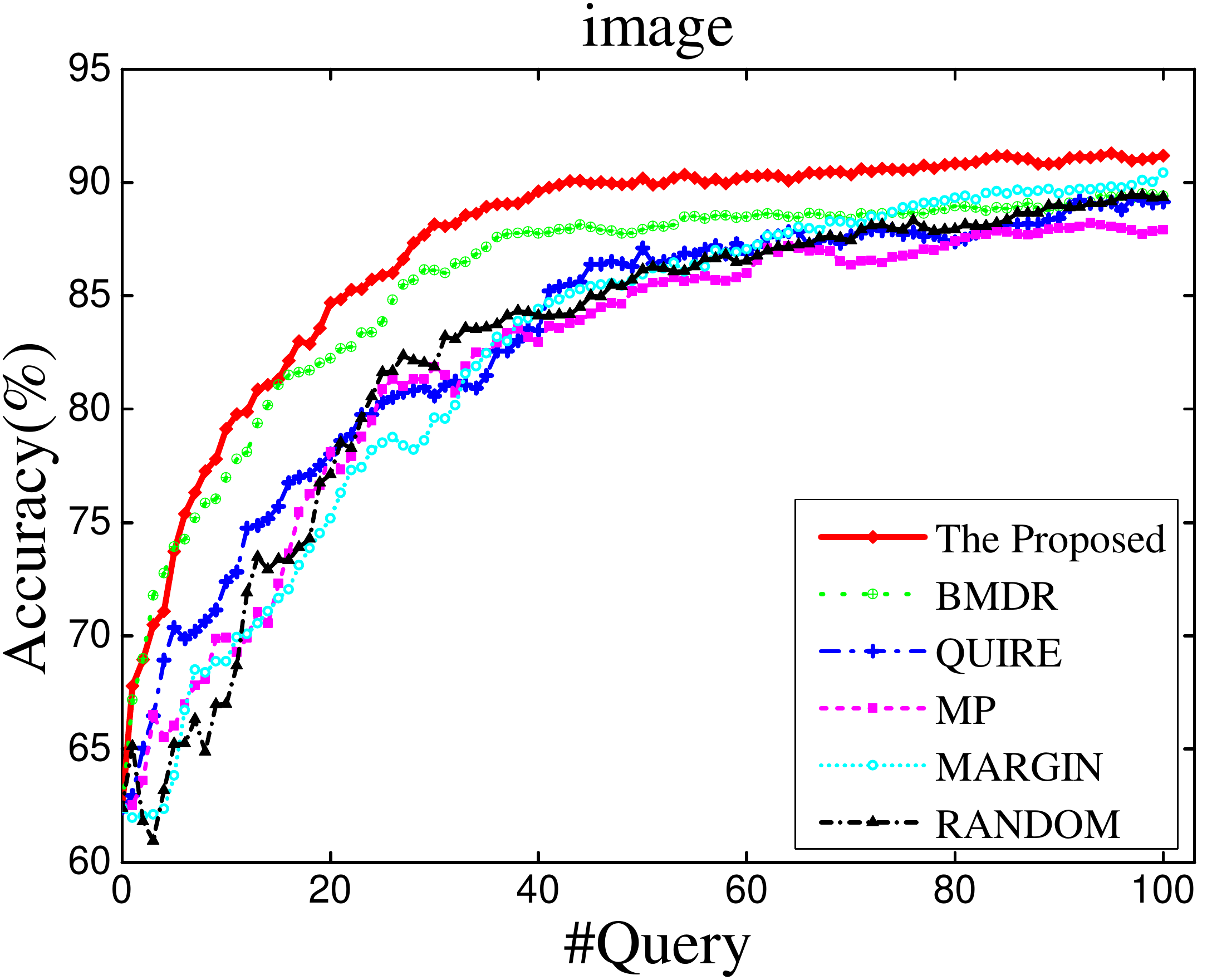, height = 1.8 in, keepaspectratio}
\epsfig{file = 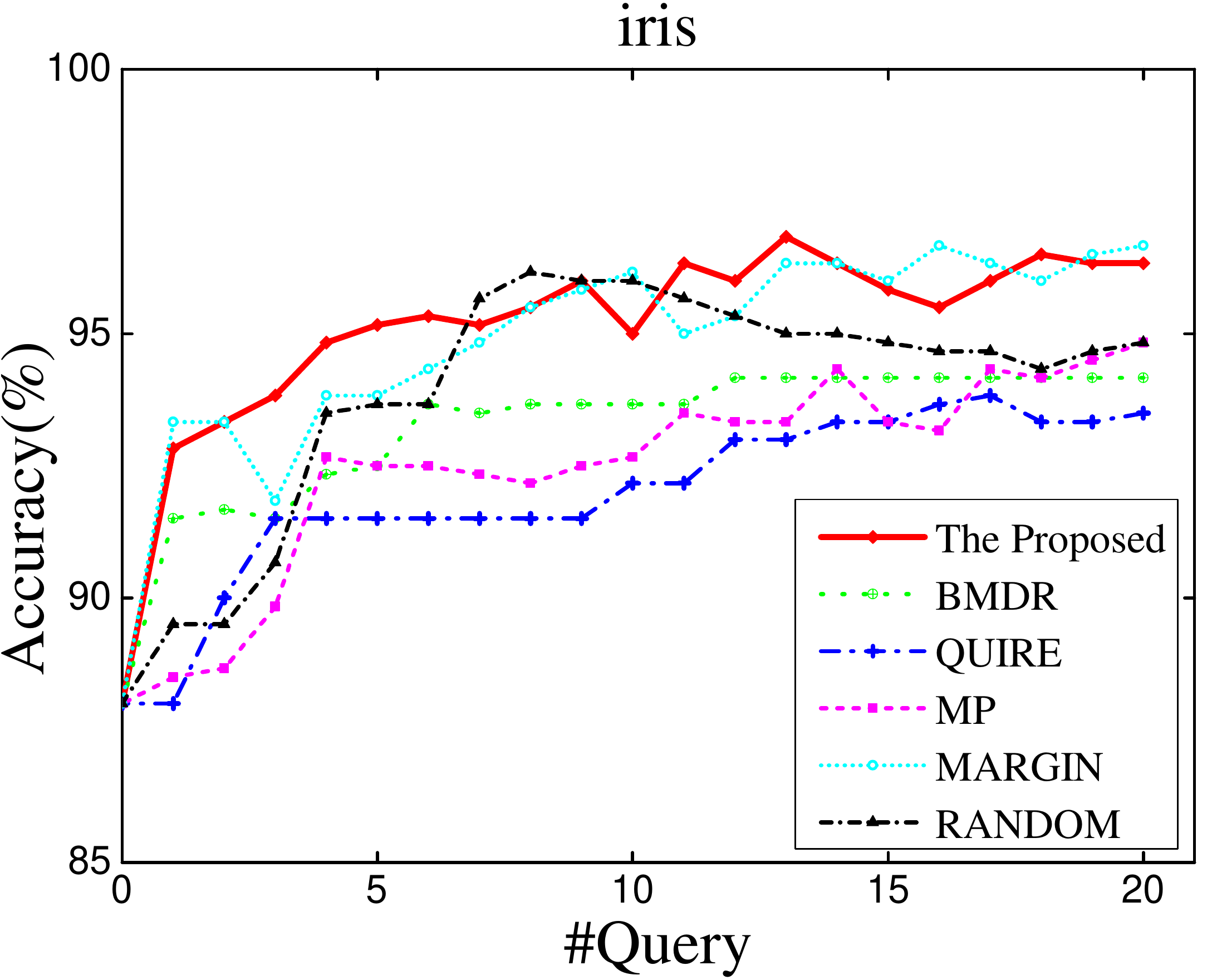, height = 1.8 in, keepaspectratio }
\epsfig{file = 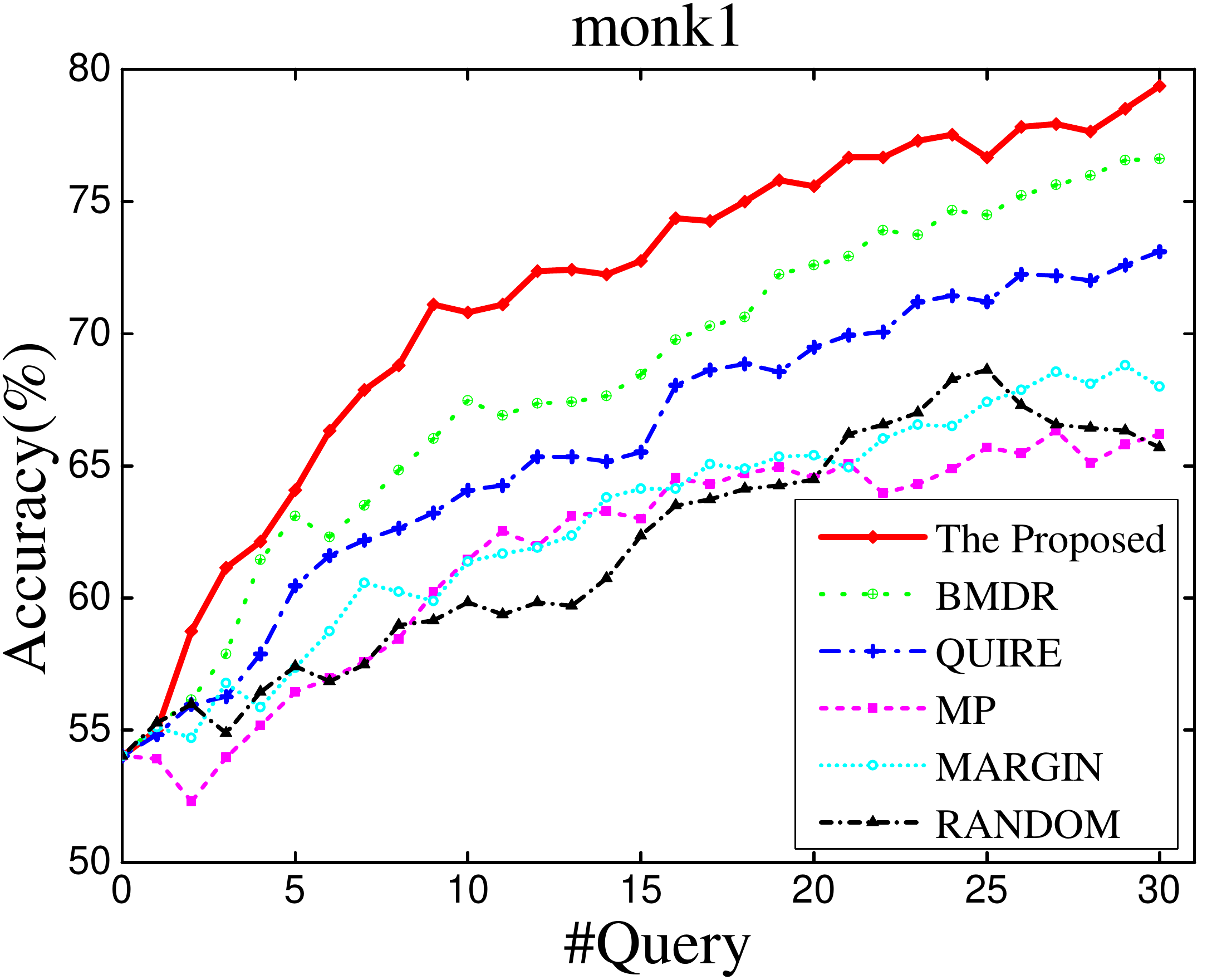, height = 1.8 in, keepaspectratio }\\
\epsfig{file = 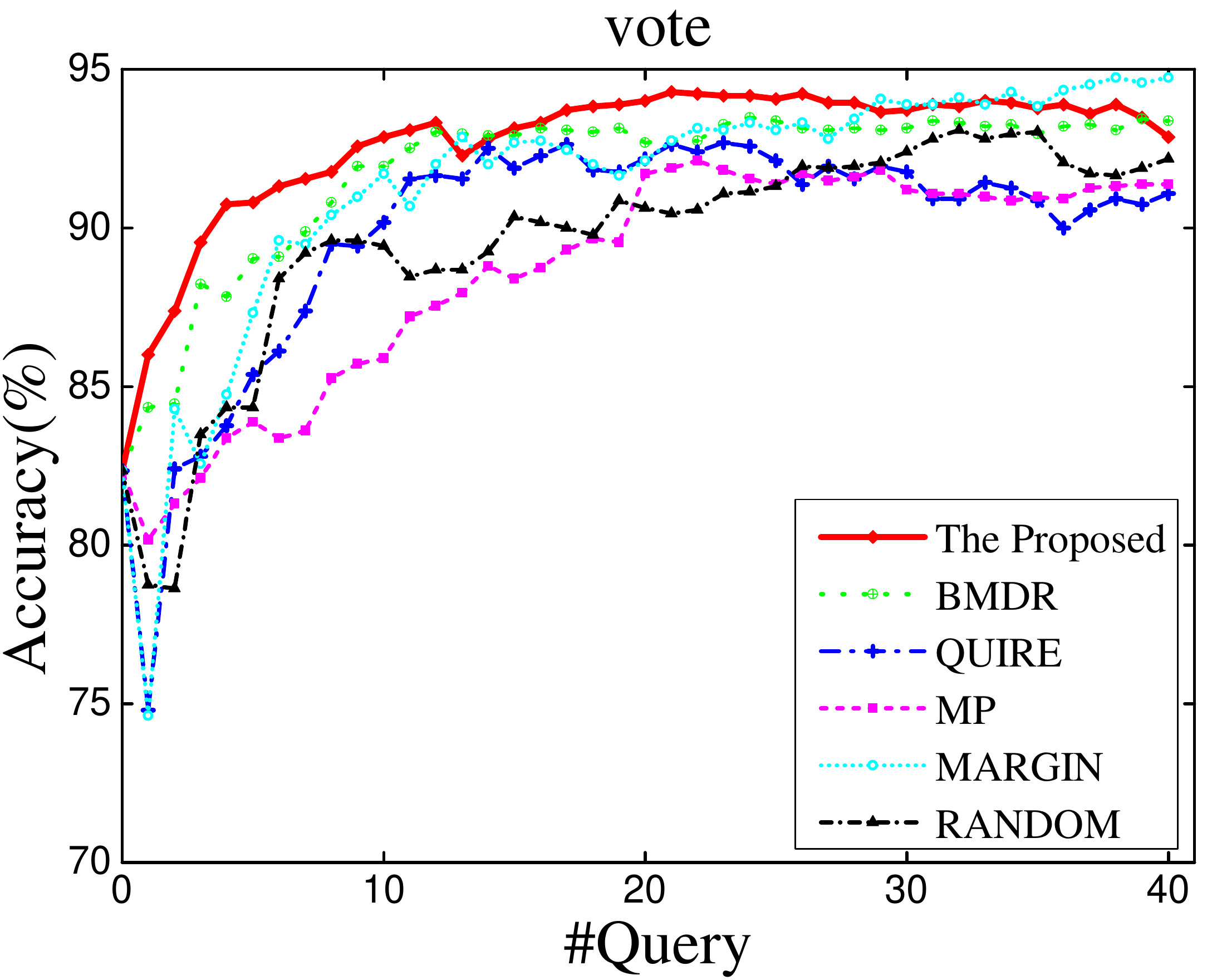, height = 1.8 in, keepaspectratio}
\epsfig{file = 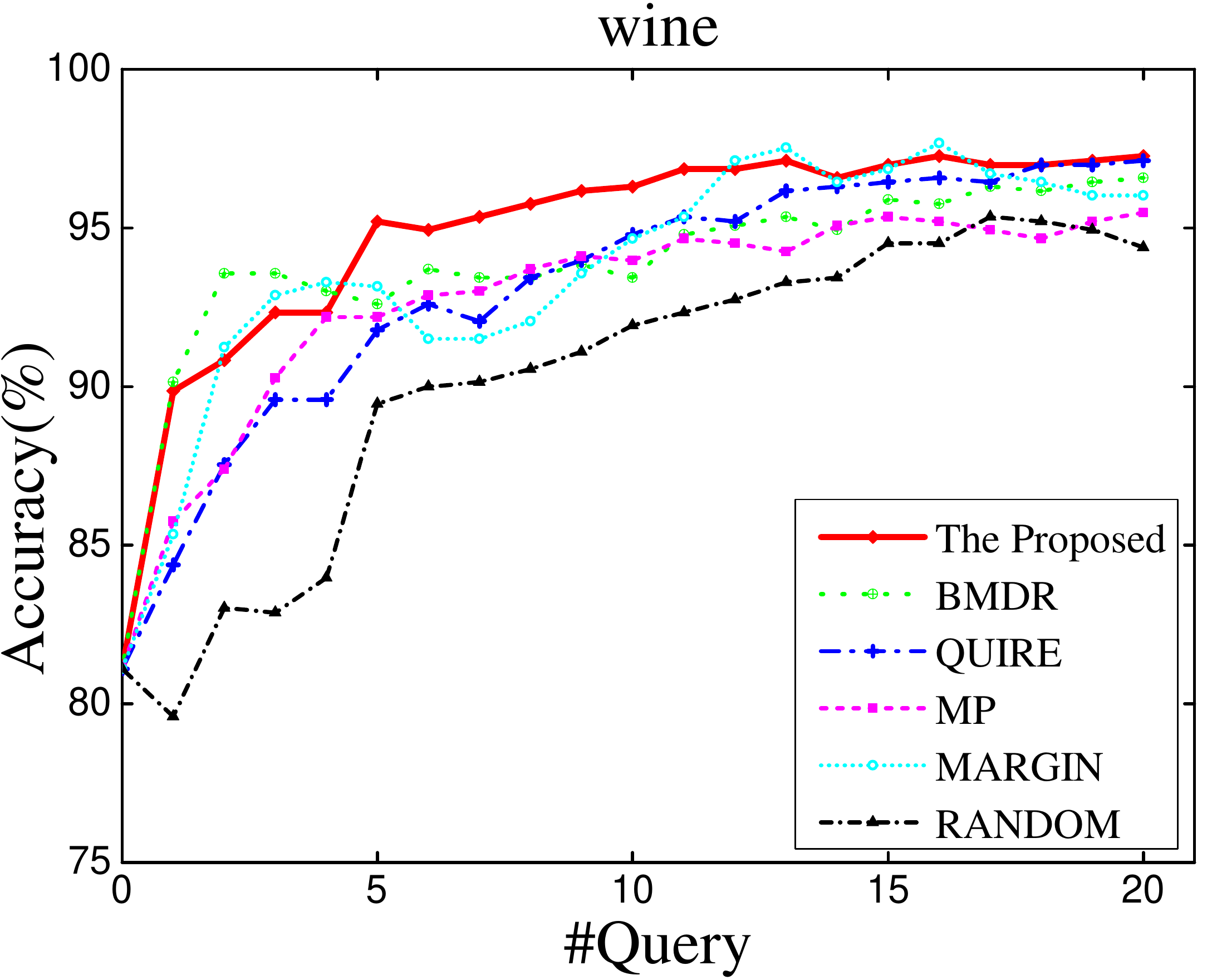, height = 1.8 in, keepaspectratio }
\epsfig{file = 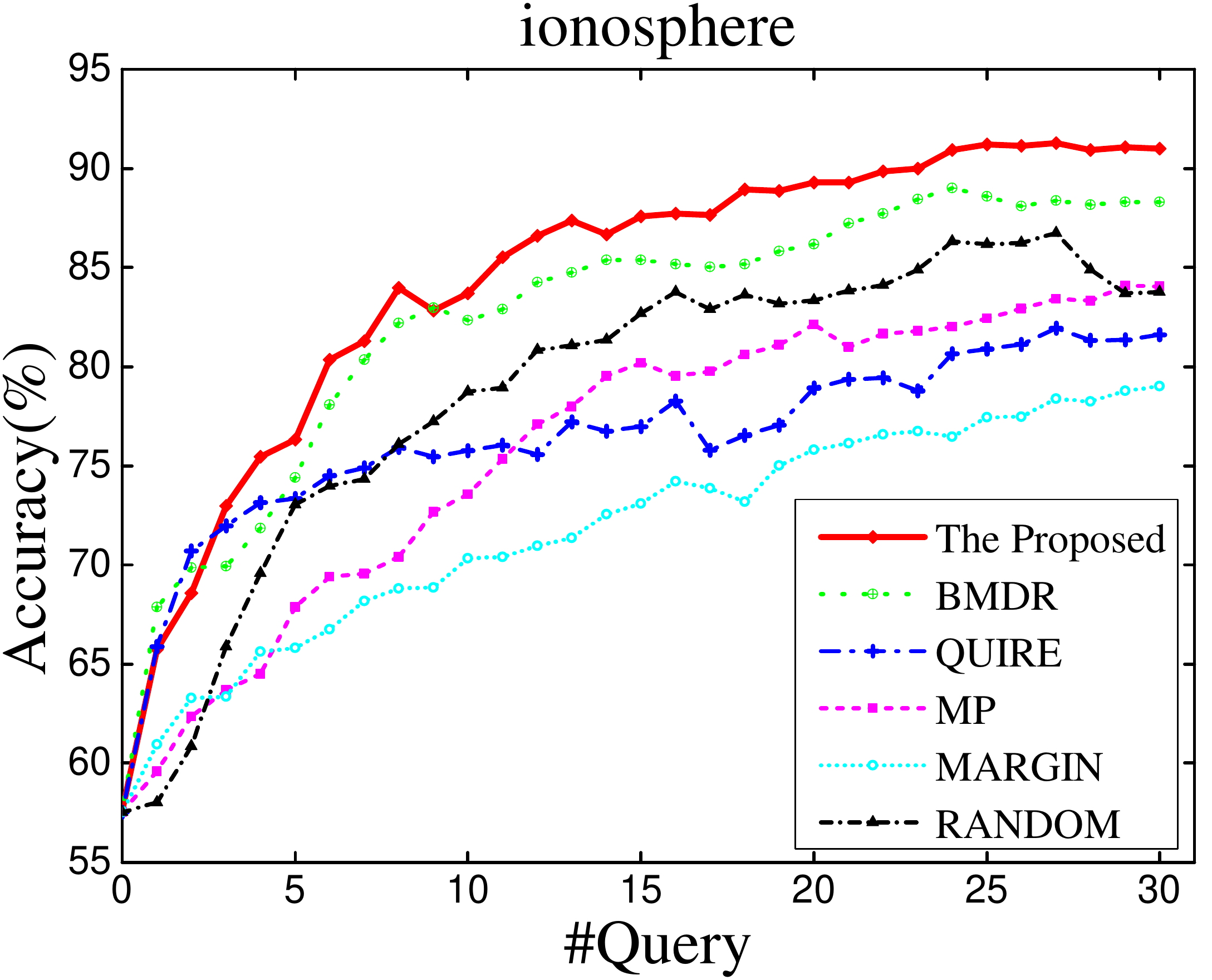, height = 1.8 in, keepaspectratio }\\
\epsfig{file = 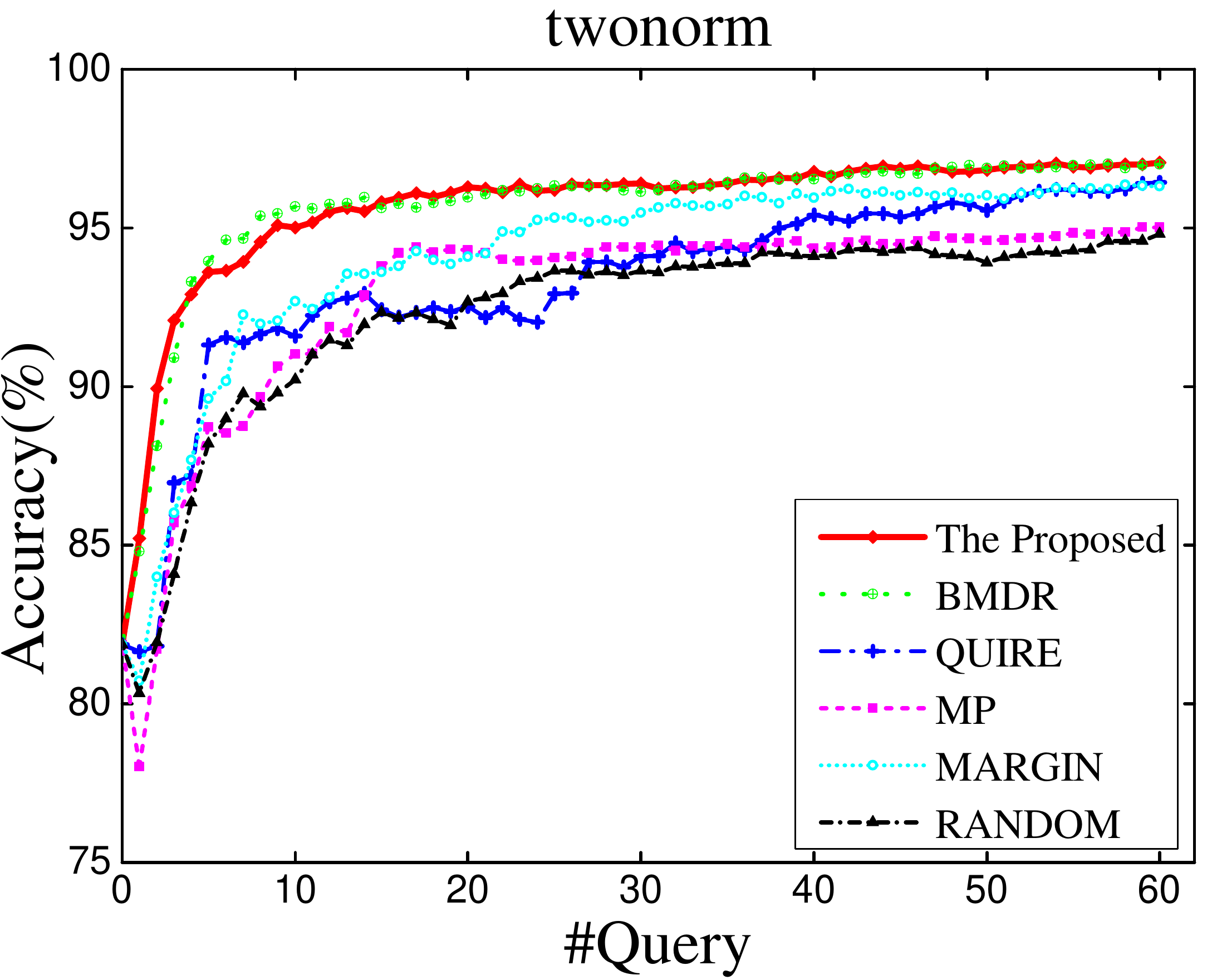,height = 1.8 in, keepaspectratio}
\epsfig{file = 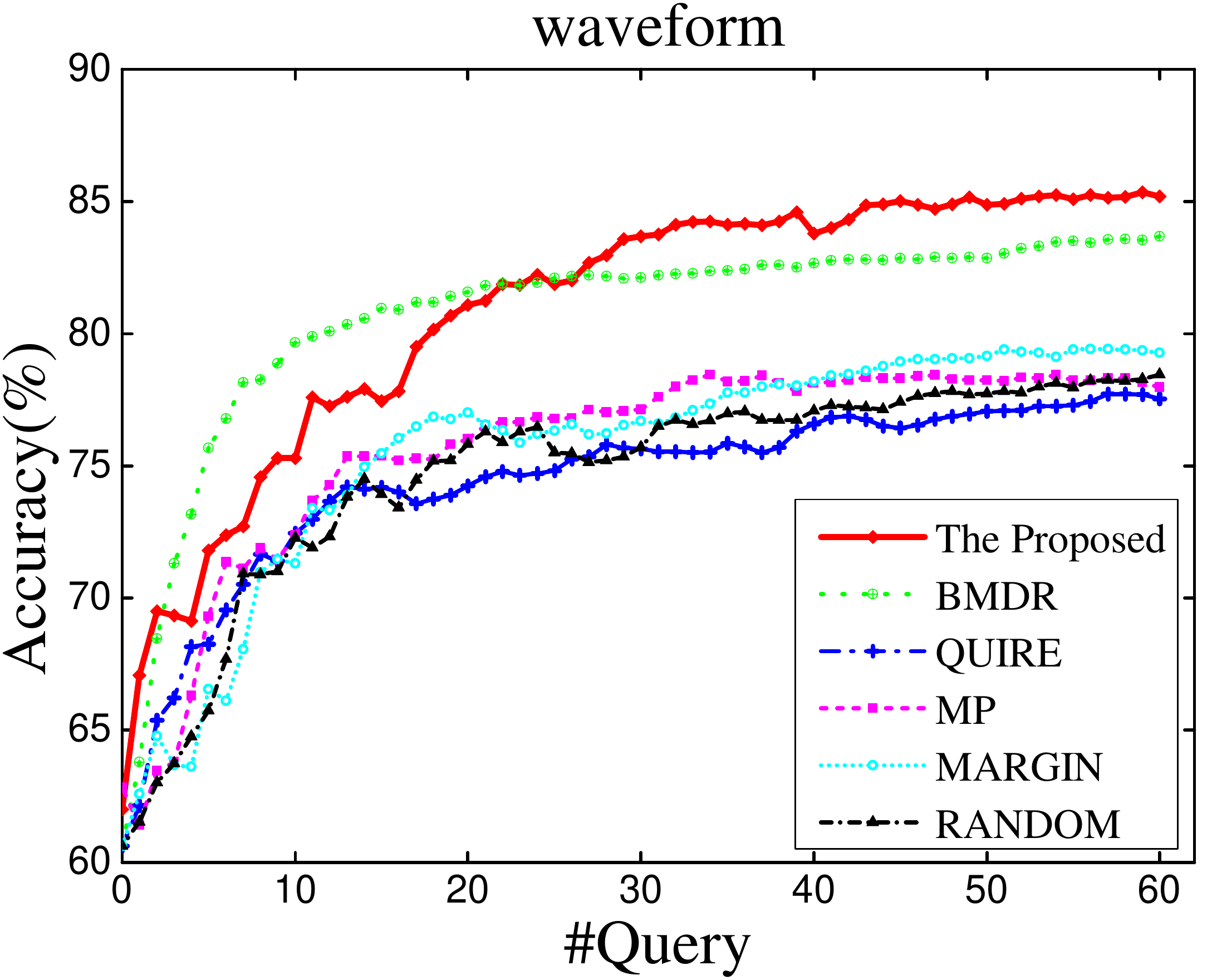, height = 1.8 in, keepaspectratio }
\epsfig{file = 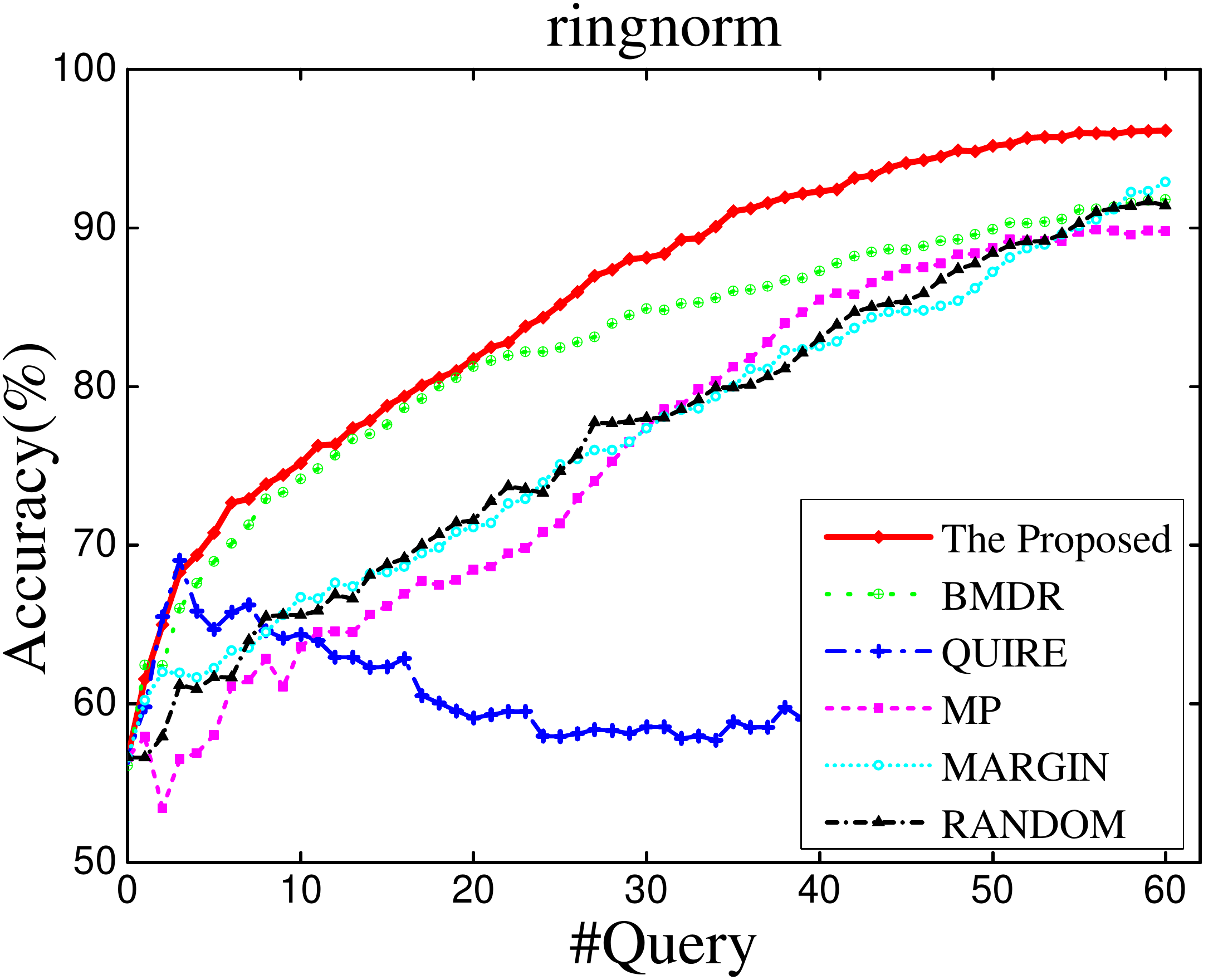, height = 1.8 in, keepaspectratio}
\caption{Comparison of different active learning methods on fifteen benchmark datasets. The curves show the learning accuracy over queries, and each curve represents the average result of 10 runs.}
\end{center}
\end{figure*}

From the results, we can observe that our proposed method yields the best performance among all the methods. The other active learning methods are not always superior to the RANDOM method in certain cases. Among the competitors, BMDR and QUIRE are two methods to query the informative and representative samples, and BMDR presents a better performance than the other competitors. It is performing well at the beginning of the learning stage. As the number of queries increases, we observe that BMDR yields decent performance, comparing with our proposed method. This phenomenon may be attributed to the fact that with a hypothetical classification model, the learned decision boundary tends to be inaccurate, and as a result, the unlabeled instances closest to the decision boundary may not be the most informative ones.  For the QUIRE, although it is also a method to query the informative and representative samples, it requires the unlabeled data to meet the semi-supervised assumption. It may be a limitation to apply the method. As to the single criterion methods, our method performs consistently better than them during the whole active learning process. The experimental results indicate that our proposed approach to directly measure the representativeness is simple but effective and comprehensive. Simultaneously, the proposed approach to measure the informativeness also contributes to the performances. By combining them together, we can select the suitable samples for classification tasks.
\begin{figure*}[ht]\label{fig2}
\begin{center}
\epsfig{file = 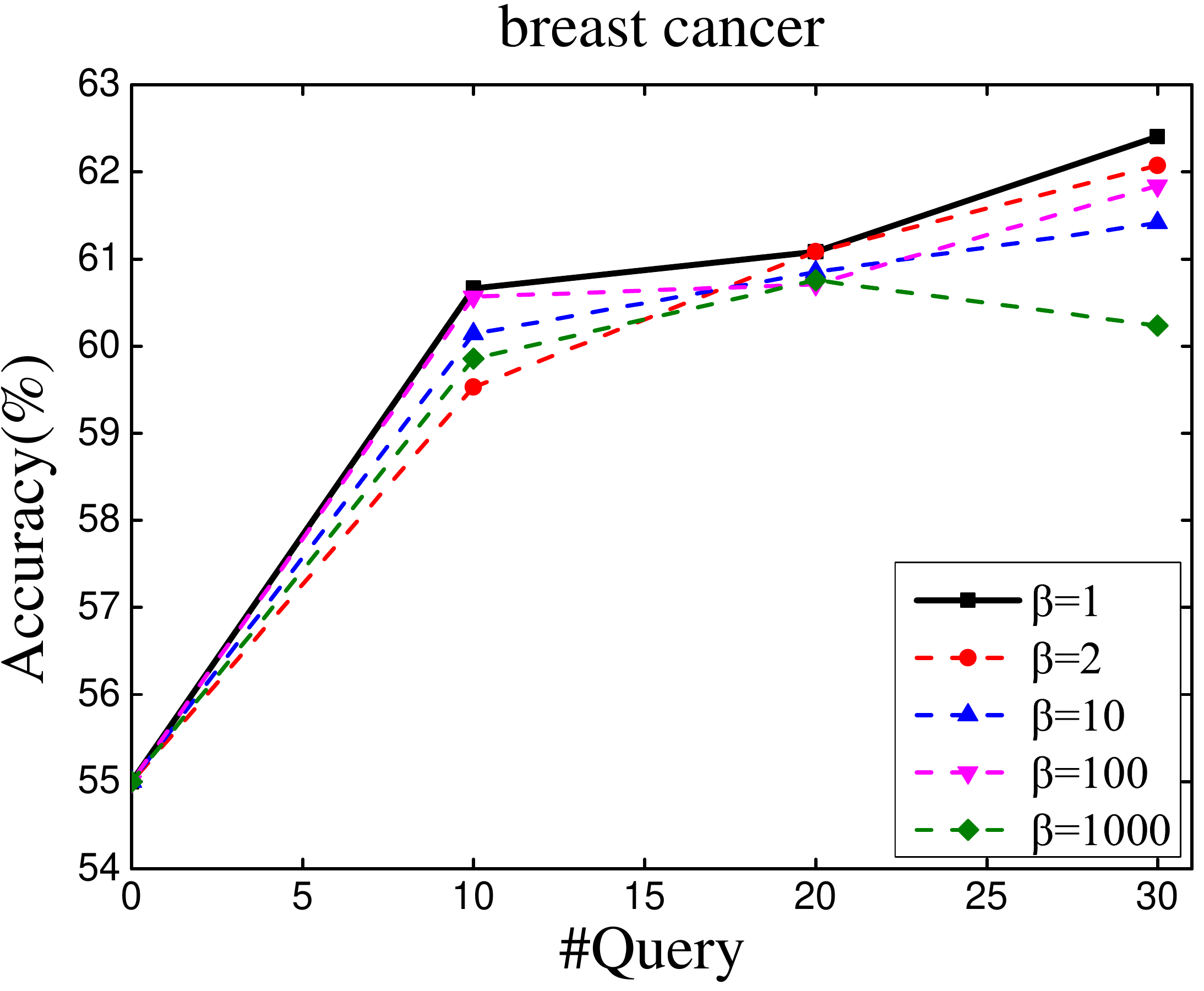, height =1.8 in, keepaspectratio}
\epsfig{file = 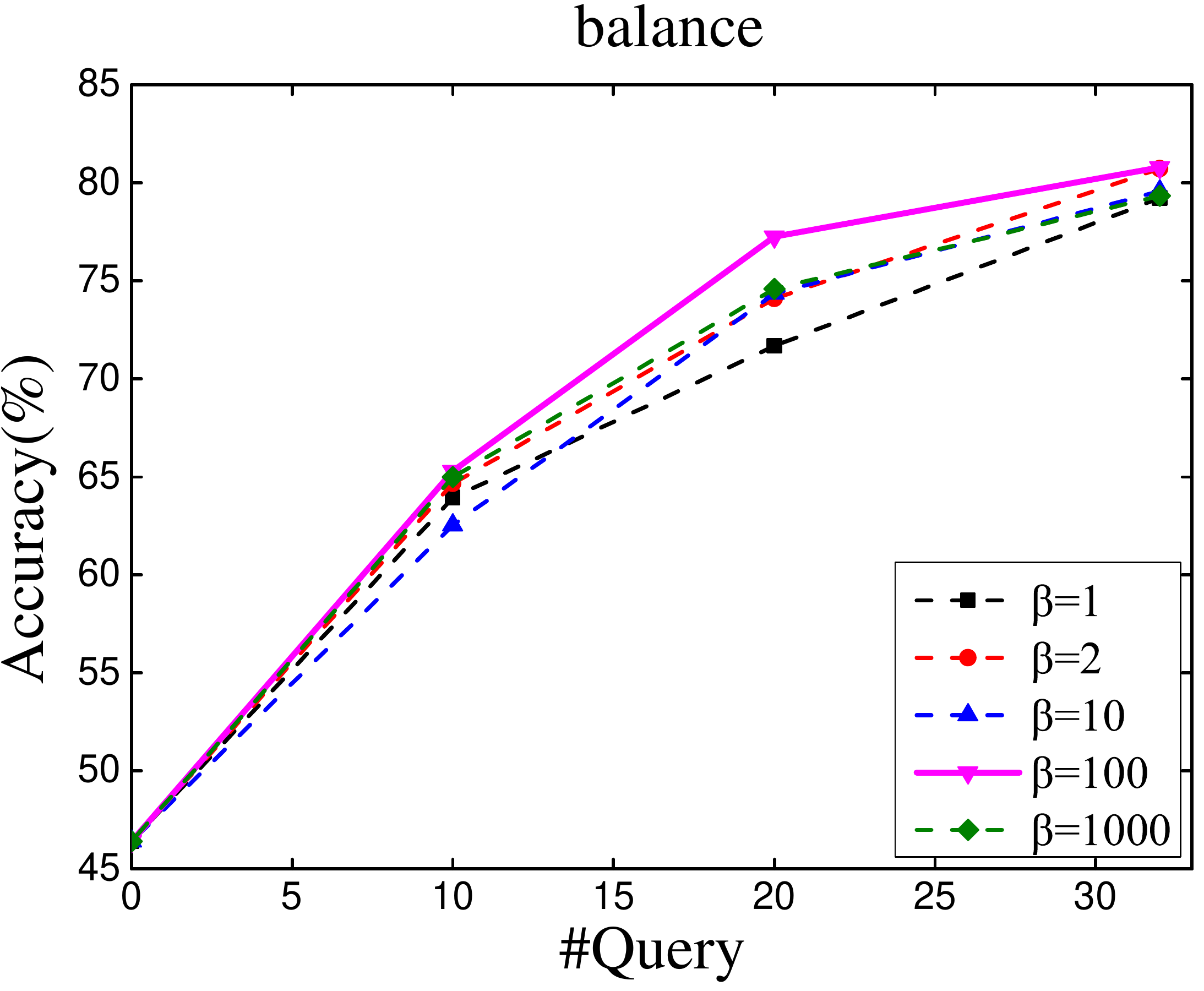, height =1.8 in, keepaspectratio}
\epsfig{file = 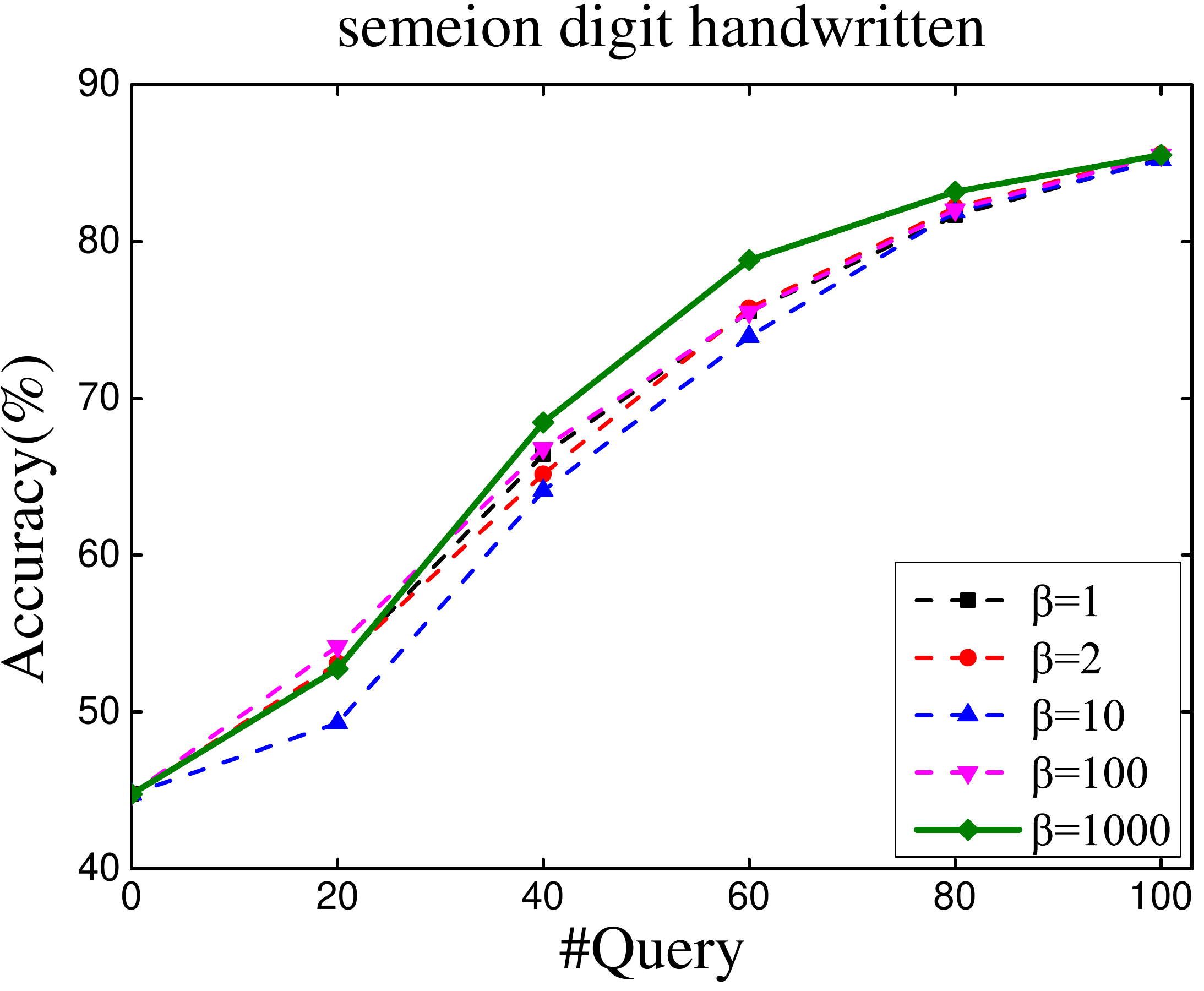, height =1.8 in, keepaspectratio}
\caption{Performance of our method with different trade-off values on three UCI benchmark datasets. Each cure represents the 10 runs average results.}
\end{center}
\end{figure*}

\section{DISCUSSION AND ANALYSIS}

In our proposed method, there is a trade-off parameter $\beta$ between the informative part and representative part in the optimization objective. In our experiments, we choose its value from a candidate set {1, 2, 10, 100, 1000}. We conduct this parameter analysis on three UCI benchmark datasets: breast cancer, balance and semeion handwritten digit. The other parameters setting are the same with the previous ones.

The performances with different $\beta$ values are shown in Figure 3. From these results, we can observe that the sensitivity of $\beta$ on three benchmark datasets are different. The performance on the breast cancer dataset is more sensitive to the $\beta$ than that on semeion handwritten digit and balance dataset. We can observe that smaller $\beta$ works better on breast cancer dastaset, while larger $\beta$ works better on semeion digit handwritten and balance dataset. In other words, the representativeness is more important to the breast cancer set, while the informativeness can better mine the data information for balance and semeion handwritten digit. The reason may be that the breast cancer data is distrbuted more densely, so the representativeness can help boost the active learning. Meanwhile, the breast cancer set just has two classes, and for each sample there are only two probabilities to measure the informativeness. Therefore, the information may not be enough. Several existing studies show that the representativeness is more useful when there is no or very few labeled data \cite{B2009,RZ2012,ZJ2013,YD2008}. However, as to semeion handwritten digit and balance, the informativeness may be dominated. This is because these data is loosely distributed. And in our method, the posterior probability is adopted. Based on the probability, we designed a new uncertain measurement, which is more suitable to measure the uncertainty of a sample. The position measure is combined into the uncertainty part, which effectively prevents the query samples bias. Besides, for these two datasets, they are multi class, so the probability information is enough to measure the amount of informativeness of a sample. This may be the reason why for semeion handwritten digit and balance perform better when the $\beta$ is larger. As the analysis above, we can infer that in our proposed method a small $\beta$ may be preferred when the dataset just has two classes; and a large $\beta$ may be recommended when the dataset has multiple classes.

Both the informativeness and representativeness are significant for active learning \cite{RZ2012,ZJ2013}. Since active learning is to iteratively select the most informative samples, and it is hard to decide which criterion is more important at each iteration, our framework provide an easy way to naturally obtain the important information in the active learning process. Meanwhile, the framework provides a principled method to design an active learning method. Through the sensitivity analysis, we can see that the distribution of dataset impacts which information is important in the active learning process. Hence, we can design more practical active learning algorithm according to the data distribution for our specific classification tasks to achieve a good performance.

\section{CONCLUSION}
In this paper, a general active learning framework is proposed by querying the informative and representative samples, which provides a systematic and direct way to measure and combine the informativeness and representativeness. Based on this framework, a novel efficient active learning algorithm is devised, which uses a modified Best-versus -Second-Best strategy to generate the informativeness measure and a radial basis function with the estimated probabilities to construct the representativeness measure, respectively. The extensive experimental results on 15 benchmark datasets corroborate that our algorithm outperforms the state-of-the-art active learning algorithms. In the future, we plan to develop more principles for measuring the representativeness and informativeness complying with specific data structures or distributions, such that more practical and specialized active learning algorithms can be produced.

\section*{Acknowledgment}
The authors would like to thank the handing editor and the anonymous reviewers for their careful reading and helpful remarks, which have contributed in improving the quality of this paper.

\ifCLASSOPTIONcaptionsoff
  \newpage
\fi

\bibliographystyle{IEEEtran}
\footnotesize
\bibliography{Ref}

%


\appendix
\begin{appendices}
{Appendix A}
\begin{proof}
According to \cite{H1984}, the proof of the Theorem \ref{th2} requires us to check two conditions. The first one is
\begin{displaymath}
\forall \theta > 0, \lim_{n\rightarrow\infty}k_n^{-2}\sum_{i=2}^{n}E\{Y_{ni}^2I(Y_{ni})>\theta k_n\} = 0
\end{displaymath}
where ${Y_{ni}=\sum_{j=1}^{i-1}H_n{(X_i,X_j)}}$, and ${k_n^2 = E(U_n^2)}$. And the second condition is
\begin{displaymath}
\lim_{n\rightarrow\infty}k_n^{-2}V_n^2 = 1
\end{displaymath}
in probability. where ${V_n^2 = \sum_{i=2}^nE\{Y_{ni}^2|X_1,...,X_{i-1}\}}$. From the two conditions, it follows that ${k_n^{-1}}$ is asymptotically normal ${N(0,1)}.$ Since
\begin{displaymath}
E(Y_{ni}^2)=\sum_{j=1}^{i-1}\sum_{k=1}^{i-1}E\{H_n(X_i,X_j)H_n(X_i,X_k)\}
\end{displaymath}
where ${2\leq i \leq n}$, then ${{k_n^2}=\sum_{i=2}^2E(Y_{ni}^2)}$. Furthermore
\begin{displaymath}
\begin{split}
&E\{H_n(X_1,X_2)H_n(X_1,X_3)H_n(X_1,X_4)H_n(X_1,X_5)\}\\
&=E\{H_n(X_1,X_2)H_n^3(X_1,X_3)\}=0
\end{split}
\end{displaymath}
and so
\begin{displaymath}
\begin{split}
& E(Y_{ni}^4)=\sum_{j=1}^{i-1}E\{H_n^4(X_i,X_j)\}\\
& +3{\sum\sum}_{1\leq j,k\leq i-1,j\neq k}E\{H_n^2(X_i,X_j)H_n^2(X_i,X_k)\}
\end{split}
\end{displaymath}
Hence
\begin{displaymath}
\begin{split}
\sum_{i=2}^nE(Y_{ni}^4)\leq const[n^2E\{H_n^4(X_1,X_2)\}\\
                        +n^3E\{H_n^2(X_1,X_2)H_n^2(X_1,X_2)\}]\\
                        \leq const.[n^3E\{H_n^4(X_1,X_2)\}]
                        \end{split}
\end{displaymath}
It now follows from the Theorem \ref{th2} that
\begin{displaymath}
\lim_{n\rightarrow\infty}\sum_{i=2}^nE(Y_{ni}^4)=0,
\end{displaymath}
which implies the condition one. We also observe that
\begin{displaymath}
\begin{split}
v_{ni}\equiv E\{Y_{ni^2}\}=\sum_{j=1}^{i-1}\sum_{k=1}^{i-1}G_n(X_j,X_k)\\
      = 2{\sum\sum}_{1\leq j \le k \leq i-1,j\neq k}G_n(X_j,X_k) + \sum_{j=1}^{i-1}G_n(X_j,X_j)
      \end{split}
\end{displaymath}
With the results in \cite{H1984} under the situations ${j_1\leq k_1}$, ${j_2\leq k_2}$, we can obtain that
\begin{displaymath}
E(V_n^4)= 2{\sum\sum}_{2\leq i \le j \leq n}E(v_{ni},v_{nj})+\sum_{i=2}^nE\{v_{ni}^2\}
\end{displaymath}
Therefore
\begin{displaymath}
\begin{split}
E(V_n^2-k_n^2)\leq const[n^4E\{G_n^2(X_1,X_2)\}+n^3E\{G_n^2(X_1,X_1)\}]\\
              \leq const[n^4E\{G_n^2(X_1,X_2)\}+n^3E\{G_n^2(X_1,X_2)\}]
              \end{split}
\end{displaymath}
It now follows from Theorem \ref{th2} that
\begin{displaymath}
k_n^{-4}E(V_n^2-k_n^2)^2\rightarrow 0,
\end{displaymath}
which proves the second condition.
\end{proof}
\end{appendices}

\begin{appendices}
{Appendix B}
\begin{proof}
Following \cite{AH1994}, we know that the two-sample discrepancy problem is used to examine $ {H_{{\sigma_1}{\sigma_2}}}$ in eq.(\ref{formula22}) under the hypothesis ${f_1 = f_2}$, and the objective is to minimize ${H_{{\sigma_1}{\sigma_2}}}$. Meanwhile, the estimators of ${ f_1}$ and ${f_2}$ are defined as in Theorem \ref{th1}. Conveniently, we assume ${\sigma_1 = \sigma_2 = \sigma}$, hence, our test is directly on  ${H_{{\sigma_1}{\sigma_2}}= H_\sigma}$. In order to assess the power of a test based on ${H_\sigma}$, the performance against a local alternative hypothesis should be ascertained. To this end, let ${f_1 = f}$  be the fixed density function, and let $g$ be a function such that ${f_2 = f + \varepsilon{g}}$ a density for all sufficiently small ${|\varepsilon|}$. Simultaneously, let ${h_\sigma}$ be the ${\alpha-level}$ critical point of the distribution of ${H_\sigma}$ under the null hypothesis ${H_0}$ that ${\varepsilon=0}$.
\begin{displaymath}
P_{H_0}(H_\sigma>{h_\sigma}) = \alpha
\end{displaymath}
Obviously, if ${H_\sigma > h_\sigma}$, ${H_0}$ is rejected. Therefore, we claim that
\begin{displaymath}
\lim_{n_1,n_2\rightarrow\infty}\alpha=0
\end{displaymath}
which is necessary if ${\hat{f}_j}$ is consistently to estimate ${f_j}$. Then, the minimum distance that can be discriminated between ${f_1}$ and ${f_2}$ is $\varepsilon= n^{-1/2}\sigma^{-p/2}$. This claim can be formalized as follows. Let ${H_1 = H_1(a)}(a\neq0)$ be the alternative hypothesis that ${\varepsilon=n^{-1/2}\sigma^{-p/2}a}$, and define
\begin{displaymath}
\hbar(a)=\lim_{n\rightarrow\infty}P_{H_1}(H_\sigma>h_\sigma)
\end{displaymath}
Actually, such a limit is well-defined, that ${\alpha < \hbar(a)<1}$ for ${0<|a|<\infty}$, and that ${\hbar(a)\rightarrow1}$ as ${|a|\rightarrow\infty}$.
This can be verified as follow. Firstly, we can observe that
\begin{displaymath}
\begin{split}
H_\sigma = &\int\{\hat{f}_1-\hat{f}_2-E_{H_1}(\hat{f}_1-\hat{f}_2)\}^2\\
      &+ 2\int\{\hat{f}_1-\hat{f}_2-E_{H_1}(\hat{f}_1-\hat{f}_2)\}E_{H_1}(\hat{f}_1-\hat{f}_2)\\
      &+ \int\{E_{H_1}(\hat{f}_1-\hat{f}_2)\}^2
\end{split}
\end{displaymath}
and ${\int\{E_{H_1}(\hat{f}_1-\hat{f}_2)\}^2}\sim\varepsilon^2\int{g}^2$. Arguing as in \cite{H1984}, if ${0< \lim_{n_1,n_2\rightarrow\infty}{n_1/n_2}<\infty}$, and ${\sigma\rightarrow0}$, ${n\sigma^p\rightarrow\infty}$, then under ${H_1}$
\begin{displaymath}
\begin{split}
& n\sigma^{(p/2)}[\int\{\hat{f}_1-\hat{f}_2-E_{H_1}(\hat{f}_1-\hat{f}_2)\}^2-\kappa_1({n_1}^{(-1/2)}\\
& +{n_2}^{(-1/2)})\sigma^{-1}],\\
&n^{(1/2)}\varepsilon^{-1}\int\{\hat{f}_1-\hat{f}_2-E_{H_1}(\hat{f}_1-\hat{f}_2)\}E_{H_1}(\hat{f}_1-\hat{f}_2)
\end{split}
\end{displaymath}
are asymptotically independent and normally distributed with zeros means and finite, nonzero variances, the latter not depending on ${a}$, where $\kappa_1=\int{K}^2$. Hence, if ${\varepsilon=n^{-\frac{1}{2}}\sigma^{-\frac{p}{2}}a}$, then under ${H_1}$
\begin{displaymath}
n\sigma^{(p/2)}[H_\sigma-\kappa_1({n_1^{-1}}+{n_2^{-1}})\sigma^{-1}-\{1+o(1)\}\varepsilon^2\int{g^2}]
\end{displaymath}
is asymptotically normally distributed with zeros means and finite, nonzero variances, the latter being an increasing function of $a$. Thus, the claims make about ${\hbar(a)}$ directly from such a result.
\end{proof}
\end{appendices}

%


\end{document}